\documentclass[lettersize,journal]{IEEEtran}
        \pdfoutput=1
\usepackage{amsfonts}
\usepackage{algorithmic}
\usepackage{algorithm}
\usepackage[caption=false,font=normalsize,labelfont=sf,textfont=sf]{subfig}
\usepackage{textcomp}
\usepackage{stfloats}
\usepackage{url}
\usepackage{verbatim}
\usepackage{cite}

\usepackage{graphicx,wrapfig,amssymb,amsfonts,amsmath,epstopdf,array}
\usepackage[dvipsnames]{xcolor}
\usepackage{tabularx}
\usepackage{array}
\usepackage{xcolor}
\usepackage{tabularray}
\usepackage{multirow}
\usepackage{ragged2e}
\usepackage{vcell}
\usepackage{soul}
\usepackage{balance}
\usepackage{caption}

\begin{document}

\title{Review on Fault Diagnosis and Fault-Tolerant Control Scheme for Robotic Manipulators: Recent Advances in AI, Machine Learning, and Digital Twin}

\author{Md Muzakkir Quamar and Ali~Nasir, Senior Member IEEE
\thanks{Md Muzakkir Quamar is with the Department
of Controls and Instrumentation Engineering, King Fahd University of Petroleum and Minerals, Dhahran, saudi Arabia
e-mail: g201806920@kfupm@edu.sa.}
\thanks{Ali Nasir is with the 1) Department
of Controls and Instrumentation Engineering and 2) Interdisciplinary Research Center for Intelligent Manufacturing and Robotics, King Fahd University of Petroleum and Minerals, Dhahran, saudi Arabia}
\thanks{Review Article}}

\maketitle

\begin{abstract}
This comprehensive review article delves into the intricate realm of fault-tolerant control (FTC) schemes tailored for robotic manipulators. Our exploration spans the historical evolution of FTC, tracing its development over time, and meticulously examines the recent breakthroughs fueled by the synergistic integration of cutting-edge technologies such as artificial intelligence (AI), machine learning (ML), and digital twin technologies (DTT). The article places a particular emphasis on the transformative influence these contemporary trends exert on the landscape of robotic manipulator control and fault tolerance.

By delving into the historical context, our aim is to provide a comprehensive understanding of the evolution of FTC schemes. This journey encompasses the transition from model-based and signal-based schemes to the role of sensors, setting the stage for an exploration of the present-day paradigm shift enabled by AI, ML, and DTT. The narrative unfolds as we dissect the intricate interplay between these advanced technologies and their applications in enhancing fault tolerance within the domain of robotic manipulators. Our review critically evaluates the impact of these advancements, shedding light on the novel methodologies, techniques, and applications that have emerged in recent times.

The overarching goal of this article is to present a comprehensive perspective on the current state of fault diagnosis and fault-tolerant control within the context of robotic manipulators, positioning our exploration within the broader framework of AI, ML, and DTT advancements. Through a meticulous examination of both historical foundations and contemporary innovations, this review significantly contributes to the existing body of knowledge, offering valuable insights for researchers, practitioners, and enthusiasts navigating the dynamic landscape of robotic manipulator control.
\end{abstract}


\IEEEpeerreviewmaketitle

\section*{Nomenclature}
\addcontentsline{toc}{section}{Nomenclature}
\begin{IEEEdescription}[\IEEEusemathlabelsep\IEEEsetlabelwidth{$V_1,V_2,V_3$}]
\item[AEs] Autoencoders.
\item[FTC] Fault tolerant control.
\item[FD] Fault diagnosis.
\item[FDI] Fault detection and identification.
\item[AI] Artificial Intelligence.
\item[DTT] Digital twin technologies.
\item[DDT] Deep digital twin.
\item[ML] Machine learning.
\item[RL] Reinforcement learning.
\item[DRL] Deep reinforcement learning.
\item[FL] Fuzzy Logic.
\item[TS] Takagi–Sugeno.
\item[SVM] Support vector machine.
\item[ANN] Artificial neural network.
\item[RNN] Recurrent neural network.
\item[CNN] Convolutional neural network.
\item[DNN] Deep neural network.
\item[RBF] Radial basis function.
\item[MPC] Model predictive control.
\item[FFT] Fast fourier transform.
\item[DWT] Discrete wavelet transform
\item[MLPN] Multi-layer perceptron network.
\item[LMPC] Lyapunov-based model predictive controller.
\item[AFT2BC] Adaptive fuzzy type-2 backstepping control.
\item[SSA] Singular spectrum analysis.
\item[HT] Hilbert Transform.
\item[LMI] linear matrix inequality.
\item[IoT] Internet of Things.
\item[RBM] Restricted boltzmann machine.
\item[DBM] Deep boltzmann machine.
\end{IEEEdescription}

\section{Introduction}

\IEEEPARstart{I}{n} an era characterized by rapid industrialization and technological advancement, robotic manipulators have emerged as indispensable tools in a wide range of industries. These mechanical arms, resembling the dexterity and precision of the human hand, have found applications in manufacturing, healthcare, logistics, and even space exploration \cite{ghodsian2023mobile}. Their ability to automate repetitive and intricate tasks has not only boosted productivity but also reduced the risks associated with dangerous and complex operations \cite{khalastchi2018fault}. However, the performance and reliability of these robotic systems are not immune to unforeseen challenges and anomalies. This is where the concept of Fault-Tolerant Control (FTC) comes into play.

The significance of robotic manipulators extends across diverse industries, revolutionizing the way processes are carried out. In manufacturing, they assemble intricate products with unparalleled precision and efficiency \cite{pedersen2016robot}. In healthcare, they assist in minimally invasive surgeries, making procedures safer and more effective \cite{noda2013impact}. In logistics, robotic arms are key players in supply chain automation, handling tasks like sorting, packing, and even delivery \cite{wahrmann}. Additionally, robotic manipulators play a pivotal role in space exploration, carrying out tasks that are too hazardous for human astronauts \cite{sreekanth2021design}. However, as robotic manipulators become increasingly integral to these industries, ensuring their reliability and safety has become paramount \cite{caccavale2003fault}. Even a minor fault or malfunction can lead to costly downtimes, product defects, and, in some cases, pose safety risks to human operators. This is where the importance of FTC becomes evident.

FTC is a field of study and practice dedicated to ensuring that robotic manipulators can continue functioning in the presence of faults or anomalies. These faults can be caused by a variety of factors, including hardware failures, environmental changes, or even external interference. The primary objective of FTC is to develop strategies that enable robotic manipulators to maintain their operations or, at the very least, carry out graceful degradation in the event of a fault \cite{jiang2012fault}. This is achieved by employing various control strategies and technologies to detect, isolate, and mitigate faults.

Fault diagnosis comprises three primary tasks: fault detection, fault isolation, and fault identification \cite{ma2011applications}. The foundational step, fault detection, involves determining whether a malfunction or fault exists in the system and identifying when it occurred. Furthermore, fault isolation seeks to pinpoint the specific location of the faulty component, while fault identification focuses on discerning the type, shape, and size of the fault. Accurately determining these factors is essential to enable the system to promptly and effectively implement fault-tolerant responses, mitigating the negative impacts caused by faulty components and ensuring normal system operation. Further, appropriate fault-tolerant control is applied to recover and reconfigure the system.

\begin{figure}[t] 
\centering
{\includegraphics[width= 90mm]{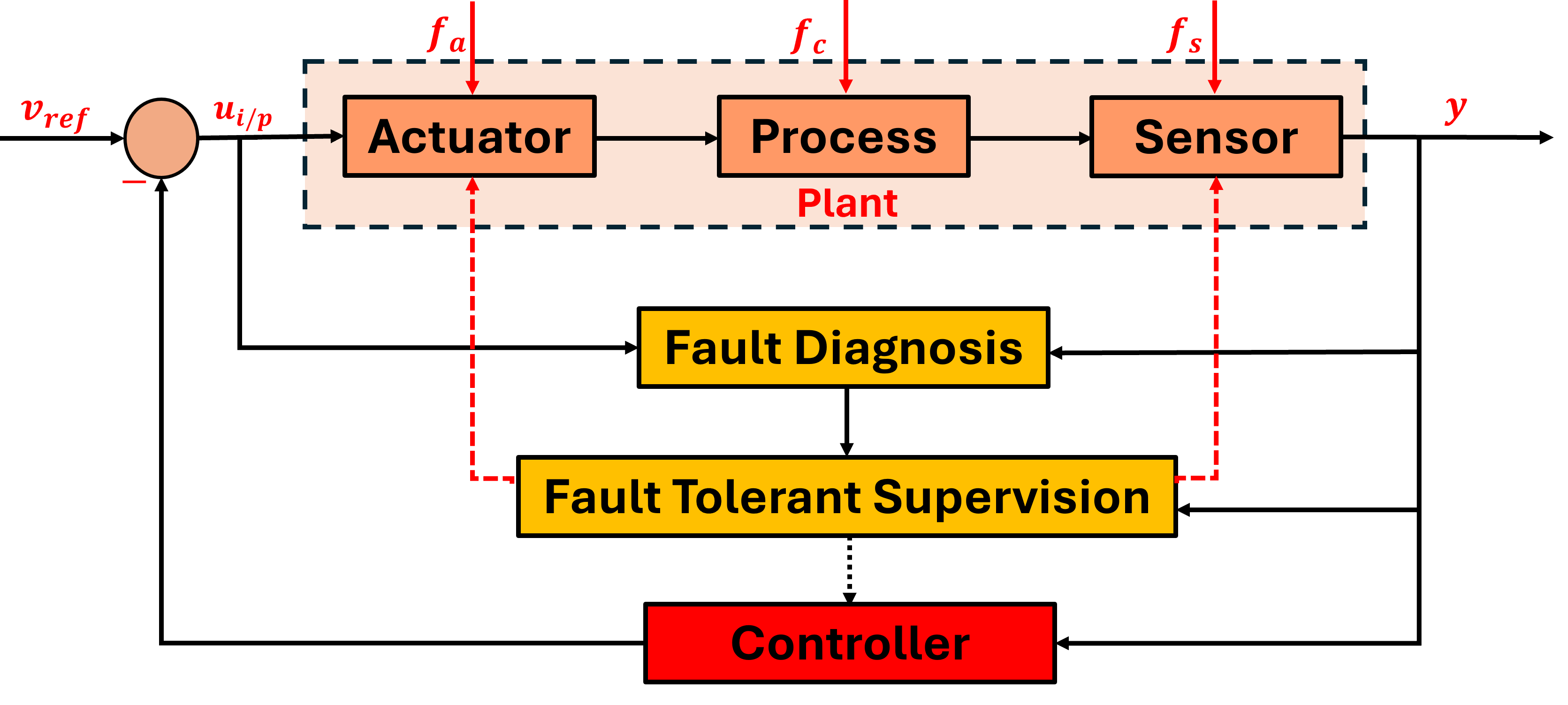}}
\caption{Schematic block structure of FD and FTC.}
\label{FTCFD}  
\end{figure}

The integration of fault diagnosis and fault-tolerant control is depicted in Figure \ref{FTCFD}. Essentially, real-time fault diagnosis plays a crucial role in detecting system faults, specifying their locations, and assessing the severity of malfunctions \cite{gao2015real}. Utilizing this valuable information, the supervision system can implement appropriate fault-tolerant measures, such as compensating for faulty signals through adjustments to actuators/sensors, fine-tuning or reconfiguring the controller, and even replacing faulty components with redundant duplicates \cite{lunze2008}. These actions are designed to accommodate or eliminate the adverse effects resulting from faults, thereby maintaining the operational integrity of the system.

As industries continue to rely on robotic manipulators for enhanced productivity and safety, understanding and implementing FTC becomes increasingly crucial. This article serves as a guide to both the historical foundations and cutting-edge innovations in this field, shedding light on the dynamic landscape of fault-tolerant control for robotic manipulators.

The main objective of this work is to provide a comprehensive overview of FTC schemes for robotic manipulators. It delves into the historical development of fault tolerance in control systems, discussing early concepts, model-based approaches, and sensor-based strategies. Moreover, it explores recent advancements that have been made possible by the integration of AI, ML, and DTT. These innovations offer more sophisticated and adaptive solutions to fault tolerance, ultimately enhancing the reliability and safety of robotic manipulators.

\subsection{Related surveys}

Over the past four decades, significant progress has been made in fault diagnosis methods, fault-tolerant control techniques, and their applications in various industries. In 2003, a comprehensive three-part survey on fault diagnosis was published in \cite{10,11,12}. A thorough overview of anomaly detection research was presented in \cite{13} with focus on identifying data patterns that deviate from expected behavior with applications in cyber security, military surveillance, and safety-critical system fault detection. The work in \cite{15,16} offered comprehensive reviews of data-driven FD methods. Additionally, in \cite{17} a concise review focusing specifically on fault detection in sensor networks is presented.

In 2015, a two-part survey on FD and FTC with detailed discussion over model-based and signal-based approaches is done in \cite{gao2015survey},  while FD with knowledge-based and hybrid approaches is discused in \cite{zhiwei2015survey}.
 A survey published in 2020, discussed fault and failure causes in control systems, analyzing the latest solutions for enhancing system resilience. It delves into recent advancements in FDI techniques and active FTC designs. Additionally, a comprehensive comparison of various FTC strategies is presented, highlighting their advantages and disadvantages. A more recent discussion on FD and FTC classification was published in 2021 \cite{fourlas2021survey}. 
 
 A comparative study between active and passive FTC scheme is discussed in detailed in \cite{jiang2012fault}. A comprehensive review of FTCS is presented in \cite{amin2019review}, covering system architecture, sensor, actuator fault types, and stability and reliability analysis. A 2020 review on FTC for robots is discussed in detail in \cite{zhang2020review}. Whereas a short review on FD for robots is presented in \cite{A2020faults}. A review on FTC for single-link flexible Manipulator System is presented in \cite{latip2012review}. Another review on fault detection and diagnosis of robotic systems was published in 2018 in \cite{2018fault}. 
 
A 2020 review on FD methods for servo motors (demagnetization, short circuits, rotor eccentricity) and drives (sensor faults, inverter faults) using signal processing and AI is presented in \cite{wu2020review}. Recent advancements are reviewed, and future trends in joint servo system fault diagnosis are discussed. While a more recent work in 2023 presented a review on AI-based fault diagnosis techniques for various industrial machines \cite{singh2023artificial}. It comprehensively analyzes key publications from the past 20 years (2000-2020), encompassing diverse AI techniques like ANN, DL, Fuzzy Logic (FL), and Support Vector Machines (SVMs).

 The most recent review of FTC for Robotic Manipulators is presented in \cite{milecki2023review}, covering publications from 2003 to 2022. The review specifically focuses on the utilization of AI techniques, including fuzzy logic, ANN, and sliding mode control, in conjunction with other control methods to achieve robust and resilient operation of robotic arms. A 2023 work with a comprehensive review on the prognostics and health management of industrial robots using DL is presented in \cite{kumar2023prognostics}. This review delves into the key areas of fault detection, diagnosis, isolation, and reconfiguration, all leveraging the power of ML, DL, and RL techniques.

\subsection{Contributions of this Survey}

My contribution to this review paper lies in addressing a crucial gap in the field of FD and FTC for robotic manipulators. While extensive research exists on both topics individually, a comprehensive review focusing specifically on robotic manipulators and incorporating recent advancements in AI, ML, DL, RL, DT does not exist in the literature as shown in Table \ref{table1}.

While a handful of studies have surveyed FD and FTC with a focus on AI and DL, they remain fragmented. This review bridges this gap by providing an integrated analysis of all these cutting-edge technologies, highlighting their potential synergy and future applications for FD and FC in robotic manipulators.

The main highlight of my work are as under:

\begin{itemize}
    \item \textbf{Provides a comprehensive and unified resource}: This review serves as a valuable reference for researchers and engineers working on FD and FTC for robotic manipulators.
    \item \textbf{Highlights the latest advancements}: The review focuses on recent developments in AI, ML, DL, RL, and DT, offering insights into their potential for enhancing FD and FTC capabilities.
    \item \textbf{Synergy Analysis}: Investigating the potential of synergistic combinations of these technologies for enhanced performance and robust fault handling.
    \item \textbf{Promotes future research}: By identifying promising research directions and potential applications, the review encourages further exploration and development in this rapidly evolving field.

\end{itemize}


\begin{table}[h!]
\centering
    \caption{ A comparision of the recent review on FD and FTC for robotic machines and manipulators using different techniques.} 

\begin{tabular}[h]{|m{2.5em}|m{1.5em}|m{1.5em}|m{1.5em}|m{1.5em}|m{1.5em}|m{3.0em}|} 
\hline
  Ref. & AI & ML & DL & RL & DT & Synergy \\
  
\hline
\cite{milecki2023review} & \checkmark & \checkmark & - & - & - & - \\
\hline
\cite{2022machine} & - & \checkmark & - & - & - &  - \\
\hline
\cite{liu2021deep} & - & - & - & \checkmark & - & -  \\
\hline
\cite{kumar2023prognostics} & - & \checkmark & \checkmark & \checkmark & - & -\\
\hline
\cite{xu2019digital} & - & - & \checkmark & - & \checkmark & - \\
\hline
This work & \checkmark & \checkmark & \checkmark & \checkmark & \checkmark & \checkmark \\
\hline
    \end{tabular}
    \label{table1}
\end{table}

\subsection{Organization}

\textbf{The structure of this article is organized to offer a detailed exploration of FTC for robotic manipulators. We begin with a historical perspective, tracking the evolution of fault tolerance strategies, and then delve into contemporary developments. We examine the pivotal role of AI and ML in fault detection, recovery, and human-in-the-loop control. We also explore the concept of DTT and its applications in predictive maintenance and real-time simulation.}

\section{Robotic Manipulator}

\begin{figure*}[t] 
\centering
{\includegraphics[width= 125mm]{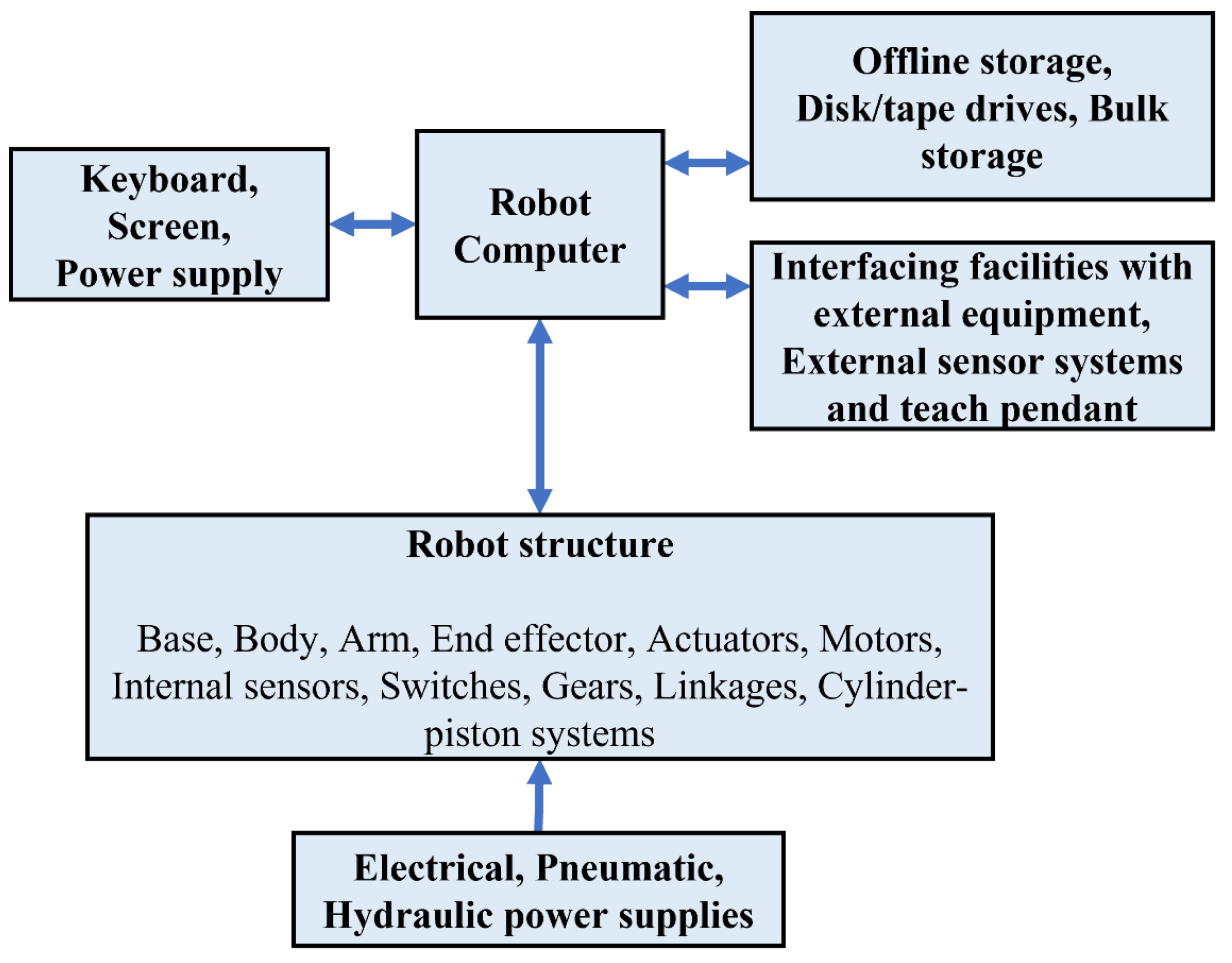}}
\caption{Simplified block structure of a robotic system \cite{kumar2023prognostics}}
\label{robot}  
\end{figure*}    

Robots have shed their initial image of clunky, industrial assistants, evolving into highly dexterous, adaptable machines capable of tackling a vast spectrum of tasks. From the cramped confines of automotive assembly lines to the delicate intricacies of minimally invasive surgery, robots now navigate environments both hazardous and intricate, exceeding the limitations of human capabilities. This remarkable transformation owes much to the advancements in mechanics, sensing, and control systems, granting robots not only dexterity but also autonomy. As a result, their applications have expanded beyond the factory floor, finding roles in fields as diverse as healthcare, underwater exploration, and even household assistance.

Within the realm of industry, robots have become indispensable partners in achieving high-precision tasks in manufacturing \cite{son2023review} and surgery \cite{das2023review}. These complex machines, often sporting intricate robotic arms and a network of sensors, are the workhorses of modern production lines and operating theatres \cite{karabegovic2016role}. However, their intricate machinery, often featuring servo motors and complex gear systems, leaves them susceptible to failure. Ensuring optimal performance and minimizing downtime for these robotic workhorses requires continuous monitoring and a deep understanding of their key components. Figure \ref{robot} provides a glimpse into its core: the power sources that fuel its actions, the control computer that orchestrates its dance, and the mechanical framework that serves as its muscular system. Within this framework lies a complex ecosystem of pneumatic, hydraulic, and electrical actuators, sensors that provide feedback, and an end-effector that serves as the robot's hand, interacting with the environment \cite{bailak2004advanced}. All these components work in unison, guided by sophisticated software and safety functions, to translate the digital commands of the control computer into precise movements \cite{yu2017achieve}.

\begin{figure}[t] 
\centering
{\includegraphics[width= 90mm]{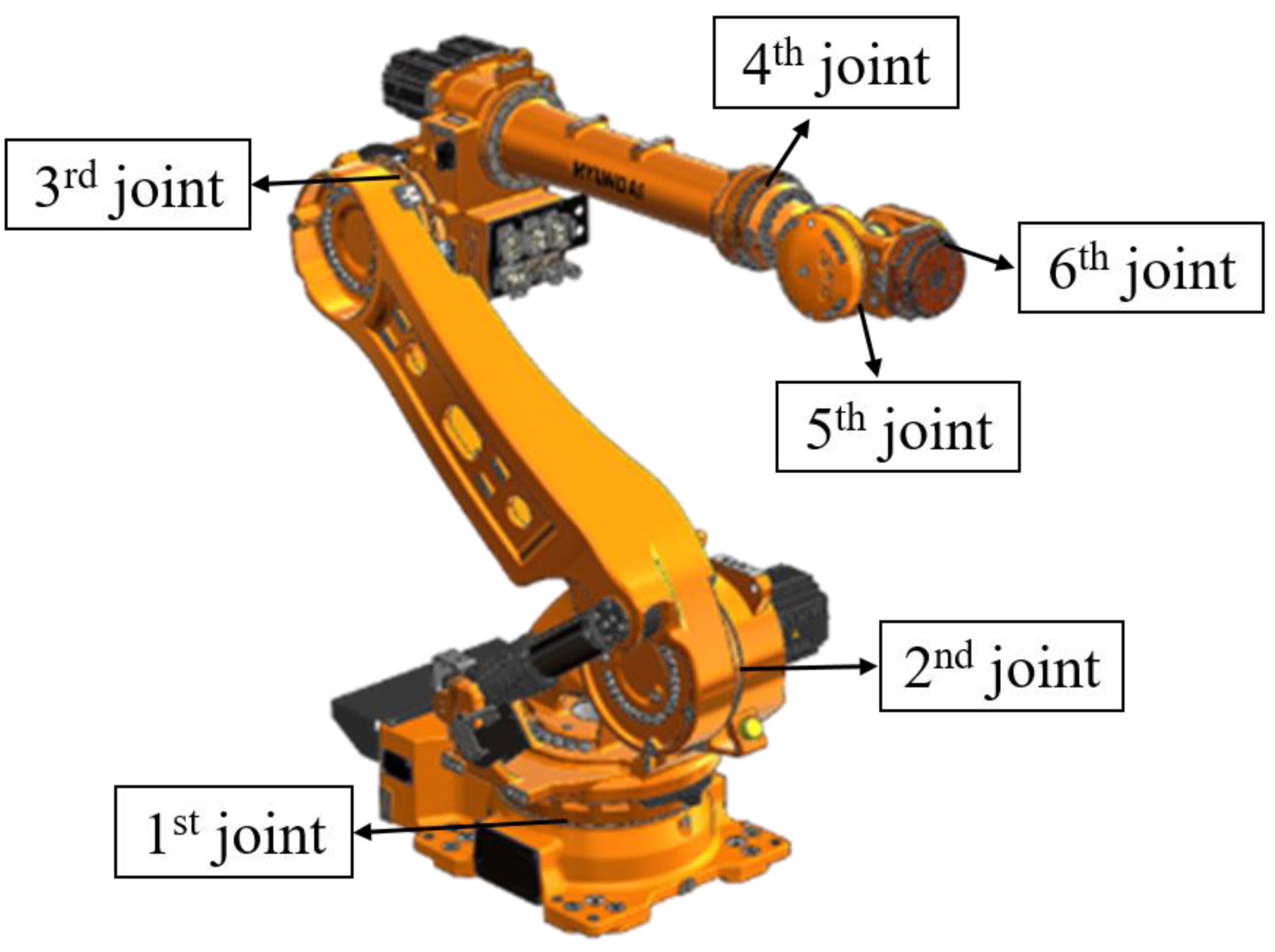}}
\caption{A typical robotic manipulator with 6-dof \cite{raouf2022mechanical}}
\label{six_dof}  
\end{figure}

However, the versatility of robots extends beyond mere movement. Programmable robots tirelessly repeat pre-defined sequences of tasks \cite{pan2012recent}, while computer-programmable robots offer the additional flexibility of remote control and complex operation \cite{sahu2022modelling}. Their structural configurations, as showcased in Figure \ref{six_dof}, can be tailored to specific needs, with simple non-servo robots handling basic tasks and six-axis robots boasting impressive dexterity and independent movement in each direction.

Despite these advancements, even the most sophisticated robots are not immune to faults \cite{wu2021fault}. Timely detection of malfunctions is crucial, as they can not only degrade performance and lead to failures but also accelerate wear and tear on delicate machinery. While a faulty robot may still produce acceptable results in the short term, its long-term efficiency and lifespan are significantly compromised \cite{khalastchi2018fault}. Therefore, understanding the types of faults that can afflict different robot components and developing robust fault detection and mitigation strategies are crucial for ensuring the smooth and reliable operation of these tireless robotic workers.

\section{Historical Development of Fault-Tolerant Control}

The fundamentals of fault tolerance in computer systems were outlined in \cite{aviziens1976fault}, which introduced the various error classifications and the use of redundancy techniques to ensure reliable computer operation. Fault modeling and prediction were also covered, as well as instances of fault-tolerant computers. In the early 1990s, \cite{stengel1991intelligent, veillette1990design}, and  \cite{patton1993robustness} published the findings of various fault-tolerant control system applications. A state-of-the-art overview of FTC methods for control applications in the late 90's was provided in \cite{patton1997fault}.

The historical development of FTC for robotic manipulators is a testament to the relentless pursuit of improving reliability and safety in automated systems \cite{chladek1990fault,wu1991fault}. A substantial body of research was dedicated to the advancement of FTC systems for application in manipulators tailored for space exploration in \cite{wu1993fault,ting1993control, chen2022review, kim2002intelligent}. The foundational principles for crafting fault-tolerant manipulators were elucidated in \cite{tesar1993design}. In this work, sophisticated modeling tools and robust assessment techniques for robotic systems were delineated. It comprehensively addressed design considerations and conducted a meticulous reliability assessment. Drawing from the outcomes of these analyses, a methodology for the design of fault-tolerant manipulators was proposed. Particular emphasis was placed on formulating recommendations for the development of a 10-degree-of-freedom robotic arm, suggesting the utilization of two actuators at each joint.

Throughout the historical development of FTC, several influential researchers and key developments have played pivotal roles in advancing the field. Notable among these is the work of John J. Craig, \cite{craig2006introduction,craig1987adaptive,craig1986introduction} whose groundbreaking research in robotics, including the development of the Denavit-Hartenberg parameters \cite{corke2007simple,prada2020application} and kinematic equations\cite{zhang2020kinematic}, laid the theoretical foundation for control strategies in robotic manipulators. These developments greatly influenced the field of FTC \cite{ben2015kinematic,krakauer1999independent}, as understanding the robotic system's dynamics and kinematics is essential for fault detection and mitigation.

Another significant milestone was the introduction of redundancy resolution methods by researchers like M.Vidyasagar \cite{vidyasagar1985reliable, chen1988optimal, vidyasagar2001randomized, 1102376}. Redundancy resolution allows manipulators to continue functioning in the presence of faults by reconfiguring the robot's motions or altering the task space while maintaining end-effector constraints \cite{wang2008international,sankaranarayanar1990path, li2012fault, li2019fault}.
As the years progressed, researchers also began incorporating sensor-based approaches into FTC \cite{gao2015survey,zug2009approach}. By integrating advanced sensors, robotic systems could monitor their state and environment in real-time, detect anomalies, and adapt control strategies accordingly. This development further enhanced the fault-tolerance capabilities of robotic manipulators.

In conclusion, FTC has grown over time by gradually moving from early ideas that relied on hardware redundancy to more advanced methods that are based on models and sensors \cite{zhang2008bibliographical,fourlas2021survey}. This evolution was driven by the collaborative efforts of influential researchers who made significant contributions to the field. These developments laid the foundation for contemporary advancements in AI, ML, and DTT, which continue to shape the landscape of fault-tolerant control for robotic manipulators.
In the subsequent sections, we will briefly discuss the emergence of model-based and sensor-based approaches.

\subsection{\textbf{Model-Based Approaches to Fault-Tolerant Control and Diagnosis}}

The journey of FTC began with early concepts rooted in hardware redundancy and basic fault detection mechanisms. In the nascent stages of robotics, engineers sought to mitigate system failures by employing redundant components, also known as hardware redundancy. These early robotics systems included spare motors, sensors, or mechanical components that could be activated if a fault was detected. While this approach provided some level of fault tolerance, it had notable limitations \cite{crestani2015enhancing, khalastchi2018fault}. One of the key limitations was that it was largely reactive.

Hardware redundancy allowed for system recovery after a fault was detected, but it did not actively prevent or predict failures. Additionally, this method added significant cost and complexity to the robotic system, as maintaining and synchronizing redundant components proved to be a challenging endeavor \cite{ hardware3, hardware5, hardware6}. However, these early concepts laid the foundation for understanding the importance of redundancy and fault detection in achieving fault tolerance \cite{hardware4}. As the field of robotics and automation progressed, researchers began to explore more sophisticated approaches \cite{hardware1, hardware2}.

The evolution of FTC saw a significant shift towards model-based approaches \cite{gao2015survey, faraz2021model, hasan2019model, habibi2019reliability, zhao2022model}. This paradigm relied on developing mathematical models of the robotic system and its environment, allowing for a deeper understanding of how the system behaves under various conditions. With these models, it became possible to predict and anticipate faults and, subsequently, design control strategies to address them proactively. Researchers in the field started working on adaptive control and robust control methodologies, which became cornerstones of model-based FTC.

Adaptive control aims to adjust control parameters in real time to adapt to changes in the system due to faults or uncertainties. This dynamic approach made it possible to maintain system performance in the face of unexpected deviations \cite{adaptive11,adaptive1, adaptive2, adaptive3, adaptive4, adaptive5, adaptive6, adaptive7, adaptive8, adaptive9, adaptive10, adaptive12, adaptive13}. On the other hand, robust control strategies were designed to provide stability and performance guarantees even in the presence of uncertainties or external disturbances. Robust control methods are well-suited to applications where the system's operating conditions may change over time, which is often the case in complex robotic manipulator environments \cite{robust1, paviglianiti2010robust, pedone2023robust, muhlegg2015l1, arici2020robust, zhang2016robust, chen2015robust, allerhand2014robust, liu2021novel, zahaf2022robust, bu2020robust}.

Another application of model-based fault tolerance is fault isolation and recovery. A model-based controller architecture for FTC is presented in \cite{niemann2010model}. The controller architecture is based on a general controller parameterization. The FTC architecture consists of two main parts, a Fault Detection and Isolation (FDI) part and a controller reconfiguration part. The theoretical basis for the architecture is given followed by an investigation of the single parts in the architecture. It is shown that the general controller parameterization is central in connection with both fault diagnosis and controller reconfiguration. Using controller parameterization creates a structured way to switch between different controllers, especially when it comes to the controller reconfiguration part. This also allows controller switching using different sets of actuators and sensors.

Another model-based FTC is studied in \cite{lao2013proactive} and a proactive fault-tolerant Lyapunov-based model predictive controller (LMPC) that can effectively deal with an incipient control actuator fault is proposed. This method for proactive fault-tolerant control combines the unique stability and robustness of LMPC with the ability to explicitly account for control actuator faults in the formulation of the MPC. The theoretical results are applied to a chemical process example, and different scenarios were simulated to demonstrate that the proposed proactive fault-tolerant model predictive control method can achieve practical stability and efficiently deal with a control actuator fault.

The work in \cite{shin1999fault} introduced a strategy for detecting faults and a robust control scheme for recovering from actuator faults in robot manipulators. While in \cite{li2013decentralised} a decentralized adaptive fuzzy control method based on the impedance approach for coordinating multiple mobile manipulators using Lyapunov synthesis is designed. In \cite{tong2010fuzzy} a decentralized adaptive control strategy for interconnected systems subject to bounded disturbances is developed. A novel decentralized hybrid adaptive fuzzy control system was devised for a class of large-scale uncertain nonlinear systems in \cite{ma2020adaptive}.

In \cite{yao2018fault} presents the structure and measurement models for a redundant parallel six-component force sensor. It examines its fault tolerance, outlines calibration models for both fault-free and fault-tolerant scenarios, and establishes an assembly strategy and testing platform for practical industrial use. A collaborative robotic manipulator with motion constraints using real-time finite-time fault-tolerant control based on recurrent neural networks (RNN) is presented in \cite{hwang2023cooperation}.

\begin{figure}[t] 
\centering
{\includegraphics[width= 90mm]{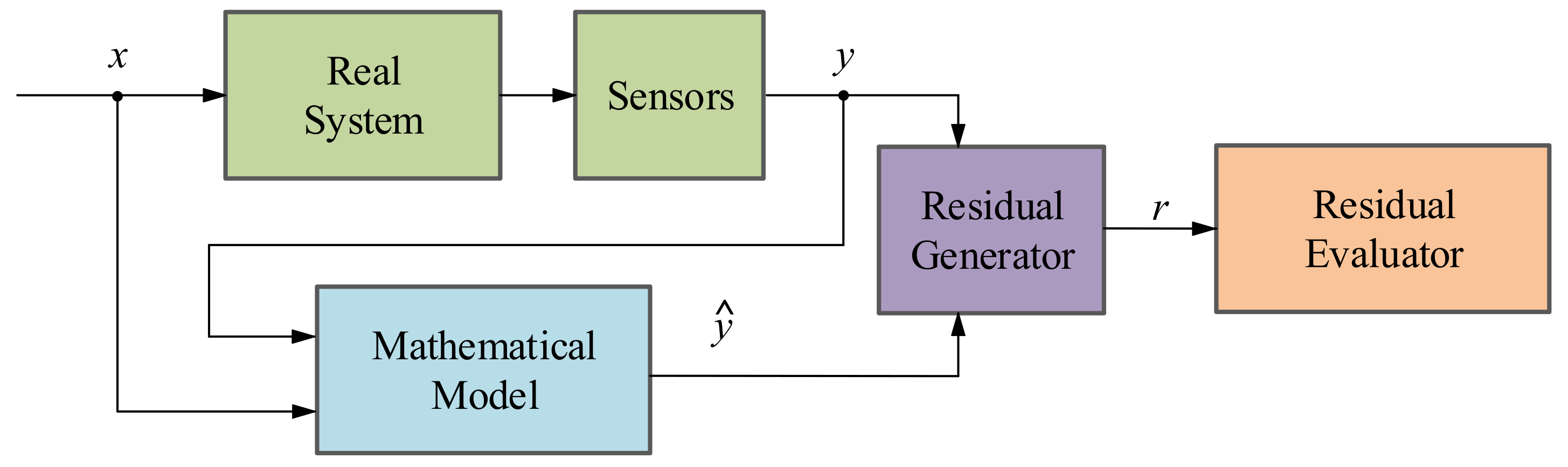}}
\caption{Simplified architecture of Model based FTC \cite{img}}
\label{model_FTC}  
\end{figure}

A novel fault detection approach centered on isolating fault signatures through filtered torque prediction error estimation is designed in \cite{dixon2000fault}. In \cite{ma2017adaptive}, authors developed an adaptive fault tolerance control scheme for two linked-wheeled mobile robots. Two fault-tolerant control strategies for robot manipulators is designed in \cite{siqueira2007fault}. The first one relies on linear parameter-varying systems, emphasizing post-fault stability when the robot halts after a fault is detected, particularly in an underactuated configuration. The second strategy is based on Markovian jump linear systems, ensuring stability even when the robot is in motion upon fault detection and accommodating all manipulator configurations within a unified model. Both strategies utilize output-feedback H$\infty$ controllers.

Although model-based approaches offer a structured and systematic means to detect, identify, and mitigate faults, they come with their own set of challenges. The accuracy and scalability of explicit input-output models for complex robotic systems, such as those found in modern manufacturing or surgical applications, remain substantial challenges. Incorporating real-time implementation while managing computational demands and adapting to diverse operating conditions adds complexity to model-based fault tolerance. Additionally, integrating model-based approaches with learning-based techniques, ensuring optimal redundancy management, and handling unknown faults are pivotal challenges that demand further exploration \cite{kumar2008model, lan2021robust, witczak2016fault}.

Looking towards the future, innovative approaches are needed to overcome these challenges and propel model-based fault tolerance and diagnosis to new heights. Advanced modeling techniques, such as the integration of physics-based and data-driven models, can enhance the accuracy of fault tolerance models. Future systems may benefit from the adoption of explainable AI techniques, ensuring transparency and interpretability in diagnostic decisions. Adaptive fault-tolerant control strategies that can dynamically adjust to novel faults and evolving conditions represent a promising avenue for research \cite{wang2015model, sarotte2020model, ding2014data}.

Furthermore, leveraging operational data for continuous model improvement, exploring edge computing for efficient real-time implementation, and designing human-centric fault-tolerant systems are critical future directions \cite{ray2022prioritized, casado2020edge, he2021challenges}. Autonomous fault recovery, where systems can automatically adapt to and recover from faults without human intervention, presents an exciting frontier for advancing model-based fault tolerance \cite{das2021explainable, awan2019adaptive}. Encouraging interdisciplinary collaboration between experts in control theory, machine learning, and domain-specific fields will be essential for developing holistic solutions that address the multifaceted challenges inherent in model-based fault tolerance and diagnosis for robotic manipulators.

\subsection{\textbf{Signal-Based Approaches to Fault-Tolerant Control and Diagnosis}}

In the realm of fault diagnosis, signal-based methods have revolutionized fault diagnosis, enabling real-time fault detection and monitoring in intricate systems. Unlike conventional model-based approaches, which depend on predefined input-output models, signal-based methods utilize measured signals to pinpoint and diagnose faults in various processes. By analyzing deviations from normal system behavior in these signals, signal-based fault tolerance and diagnosis techniques extract features indicative of potential faults. Diagnostic decisions are then made by analyzing these symptoms and drawing upon prior knowledge of healthy system behaviors \cite{gangsar2020signal, dai2019signal}.

A crucial aspect of signal-based fault diagnosis is the extraction of relevant features from measured signals using various techniques such as time-domain analysis, frequency-domain analysis, and time-frequency analysis \cite{mogal2014brief}. Time-domain features, such as mean and standard deviation, provide insights into overall signal behavior. Frequency-domain analysis, employing methods like the Fourier transform, identifies specific frequency components associated with faults \cite{li2019review, niu2023motor}. Time-frequency analysis, achieved through wavelet transforms, offers a detailed representation in both time and frequency domains \cite{islam2023designing, bai2023application}. For robotic manipulators, this involves scrutinizing signals related to joint positions, velocities, torques, and sensor feedback. Features extracted from these signals serve as indicators of potential faults, offering insights into deviations from expected behavior \cite{shi2023review,A2020faults}. 

Symptom analysis systematically examines these extracted features to identify fault patterns, relying on a profound understanding of relationships between symptoms and fault presence \cite{fourlas2021survey}. In the case of robotic manipulators, a thorough analysis of the extracted characteristics is necessary in order to discern patterns that may serve as indicators of faults. This procedure necessitates an in-depth understanding of the interconnections between certain symptoms and the probable existence of anomalies \cite{he2021challenges}. Symptom analysis serves as a fundamental component of efficient fault detection systems, including various manifestations such as deviations in joint trajectories, an unexpected increase in motor currents, or irregularities in end-effector positions \cite{kirchner2013rosha, khalastchi2018fault}.

In robotic manipulators, joint behaviors serve as vital indicators of system health. Deviations from expected trajectories, whether due to wear and tear or mechanical failures, are swiftly identified through the analysis of measured signals. Signal-based fault diagnosis enables real-time monitoring of joint positions and velocities. Following signal capture, various signal processing techniques can be employed to extract sensitive features for fault diagnosis \cite{jaber2017signal, goyal2021applications}. Industrial robots operate under diverse conditions, including varying joint speeds, loading, and articulation. The inherent non-linear dynamics of robot motion, characterized by acceleration, constant speed, and deceleration, introduce challenges for conventional signal analysis methods like fast fourier transform (FFT). Utilizing joint time-frequency techniques, such as discrete wavelet transform (DWT) \cite{yan2014wavelets} or short-time Fourier transform \cite{burriel2017short}, becomes essential for accurate robot fault diagnosis, providing precise information about energy distribution over frequency bands \cite{ sawicki2009, al2011vibration}. These techniques ensure the continued precision and accuracy of manipulator movements, crucial for applications ranging from manufacturing to surgical robotics.

Figure 5
\begin{figure}[t] 
\centering
{\includegraphics[width= 90mm]{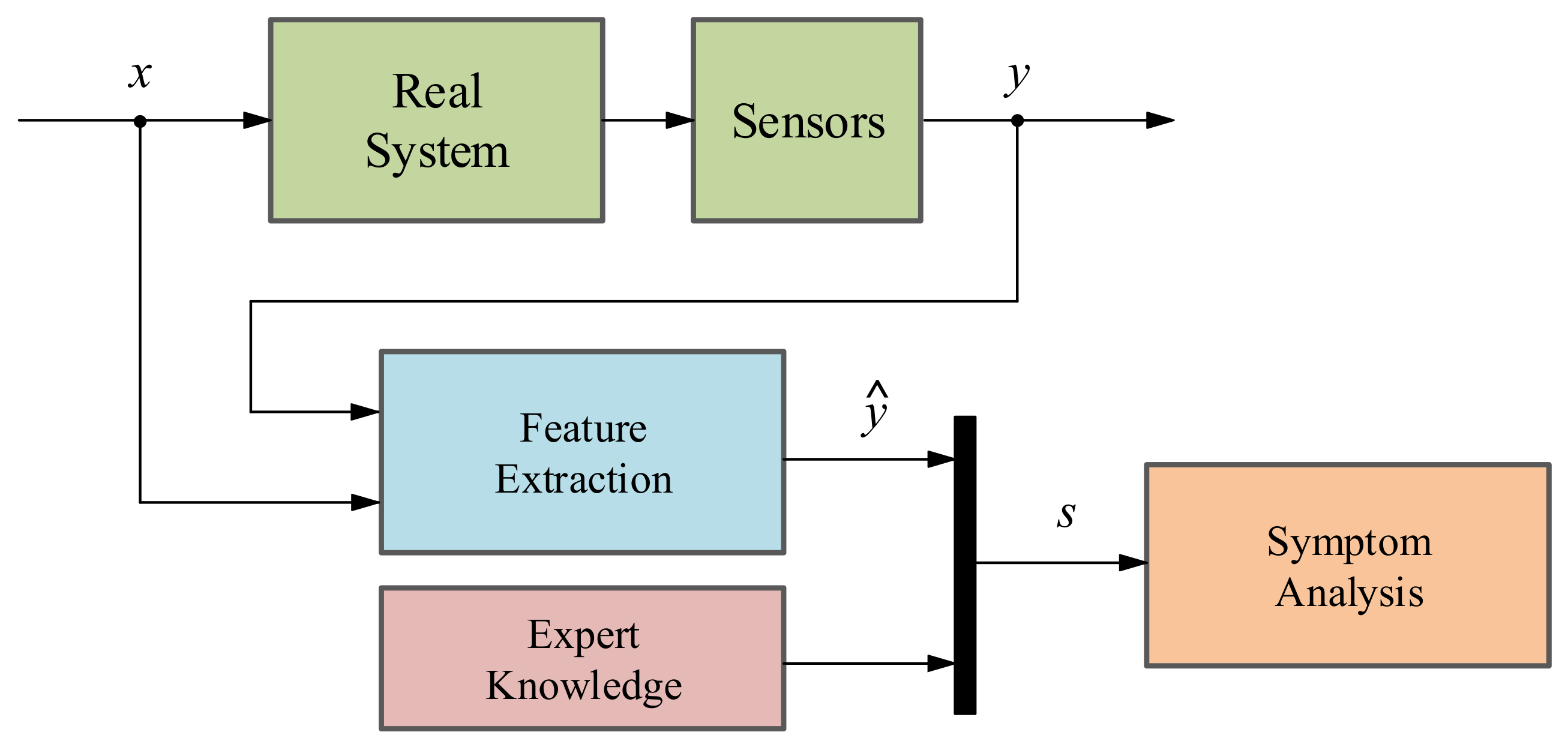}}
\caption{Simplified architecture of Signal-based FTC \cite{img}}
\label{signal_FTC}  
\end{figure}
Figure 5 ends

In \cite{jaber2014}, a fault detection and diagnosis study is conducted on a PUMA 560 industrial robot with six degrees of freedom and three links. The study focuses on replacing one of the motors, specifically at joint two, with an abnormal one. For the detection of vibration signals, three accelerometer sensors capturing (X, Y, and Z) dimensions are employed in this joint. These signals are then transmitted using a 16-bit NI data acquisition card to a PC for processing in an environment that utilizes LabVIEW and Matlab. Signal conditioning and noise removal are initially performed using the DWT. Subsequently, the signals undergo further modification through the wavelet transform to distinguish between high and low frequencies for multi-resolution analysis.

In \cite{jaber2016fault}, an intelligent system is introduced for monitoring and diagnosing gear faults in the PUMA 560 robot arm joint. The method employs the DWT in conjunction with a multi-layer perceptron network (MLPN) for fault diagnosis and classification. A significant neural network level ensures a 100

The study in \cite{alobaidy2022} presents fault detection and diagnosis for two robot arms, where faults at the joints induce oscillations at the arm's tip. Acceleration in multiple directions is analyzed to extract fault features, and simulations for planar and space robots are demonstrated. Two signal processing techniques for fault detection, namely DWT and the improved Slantlet transform is utilised. Fault allocation and classification are performed using a multi-layer perceptron artificial neural network. Results show that the Slantlet transform with the multi-layer perceptron network outperforms DWT technique, exhibiting superior performance, shorter processing time, and higher accuracy.

The work in \cite{algburi2019}, demonstrates the efficient use of encoder signals, processed with Singular Spectrum Analysis (SSA) and Hilbert Transform (HT), for observing the energy performance and health of industrial robot systems. Extracting instantaneous amplitudes (IA) and instantaneous frequencies (IF) using SSA and HT enhances the detection of weak oscillations and residuals, crucial for health evaluation and fault detection in industrial robots. The system's efficiency is verified through numerical simulation and experimental data, showcasing its potential as a promising tool for health monitoring in industrial robots, surpassing traditional vibration-based schemes.

A Gaussian mixture model-based unsupervised fault detection framework for efficient fault detection in industrial robots using current signals is suggested in \cite{cheng2019high}. Initial signal preprocessing cleans raw current signals, and motion-insensitive fault features reflecting industrial robot deterioration are then extracted. These features are input into unsupervised learning algorithms, demonstrating effective fault detection with high accuracy, as validated by experimental data from industrial robot systems.

Predictive maintenance is a crucial aspect of Industry 4.0. A predictive maintenance fault detection method is proposed in \cite{bonci2019predictive} using motor current signal analysis for industrial robots. Various signal processing techniques are utilized for the motor current generated by the Cartesian robots, which reveals a promising fault index for detecting transmission system functionality. Preliminary results show encouraging outcomes compared to traditional spectral analysis. The real-time capabilities of signal-based fault tolerance enable the implementation of predictive maintenance strategies for robotic manipulators. By continuously monitoring signals related to joint behaviors, torques, and sensor feedback, the diagnostic system can predict potential failures before they escalate. This proactive approach minimizes downtime, reduces maintenance costs, and enhances the overall lifespan of robotic manipulators \cite{divya2023review}.

Signal-based fault tolerance is crucial for safe human-robot collaboration. By monitoring force signals, robots can detect anomalies and take action to protect humans. Challenges exist, including interpreting complex signals, requiring diverse datasets, and integrating these methods into real-world robots. Future research needs to focus on explainable AI, unsupervised learning for fault detection, and novel signal processing methods.

\section{Role of Sensors in Fault-Tolerant Control and Diagnosis}


Modern equipment and machinery including robotic manipulators integrate an array of sensors, including encoders, accelerometers, and force/torque sensors. Model and signal-based fault tolerance excels in the realm of \textbf{sensor fusion}, where data from multiple sensors is amalgamated for comprehensive fault detection. Whether it's detecting irregularities in joint movements or identifying anomalies in end-effector forces, sensor fusion enhances the diagnostic system's ability to discern faults with precision. Sensor-based fault tolerance and diagnosis have become increasingly pivotal in the field of FTC \cite{kang2020sensor, mi2017fault}. They provide real-time data that enables the detection and diagnosis of faults, ultimately enhancing the overall fault tolerance of these systems \cite{baioumy2021fault, ali2021fault, truong2021novel, makni2020robust}. In this section, we delve into the role of sensors in FTC and their evolution, focusing on advancements in sensor technology and how sensor data is integrated into control strategies to improve fault tolerance.

Sensors serve as the sensory organs of robotic manipulators, providing critical information about the manipulator's state and its environment. These sensors can range from simple encoders that measure joint angles to more complex devices like accelerometers, force/torque sensors, and vision systems \cite{baioumy2021fault}. The importance of sensors in FTC lies in their ability to detect deviations from the expected behavior of the robotic system, which may result from component failures, wear and tear, or external disturbances \cite{luong2020smart, li2020recent}. When a fault occurs, sensors act as the system's "nervous system," collecting data on positions, velocities, forces, and other relevant parameters. This data is then analyzed to identify anomalies or discrepancies between the expected and actual states of the system.

The prevailing literature predominantly emphasizes actuator-based fault diagnosis, yet it is imperative to underscore the importance of implementing sensor-based fault detection and diagnosis. Sensor malfunctions can exert a profound impact on manipulator control, as inaccuracies in the measured joint angles or angular accelerations can potentially result in chaotic behavior or damage to the robotic application \cite{paviglianiti2010robust}. Different control strategies are applied to detect diagnose and reconfigure the sensor faults in a robotic manipulator. Higher order sliding-mode observer is employed in \cite{capisani2012manipulator} for the purpose of diagnosing sensor faults in the manipulator. The work in \cite{van2018adaptive,khireddine2013fault} introduces novel fault detection methods that utilize sliding-mode observers and fault tolerant control to identify faults in robot components. A fault-tolerant control system employing a sliding mode controller and a linear extended state observer for a 4-degree-of-freedom robotic manipulator is developed in \cite{tran2022fault}.

A fault-tolerant multi-sensor switching strategy for redundant sensor-equipped robot manipulators is discussed in \cite{mi2017fault}. It employs a linear-parameter-varying model to describe the robot's nonlinear dynamics and employs a zonotopic set-based robust fault detection (FD) mechanism to monitor the system. The FD mechanism checks sensors for faults and activates a switching controller to choose the best estimation from healthy sensors for control. If a fault is detected, the strategy switches to healthy sensors for control. Fault identification and isolation for robot manipulators with sensor malfunction utilizing an adaptive observer approach is presented in \cite{zeng2015adaptive}.

Sensor fault detection and compensation for robotic manipulators with adaptive observer and a terminal sliding mode control law based on a second-order integral sliding surface is discussed in \cite{ebrahimi2023sensor}. This method allows sensor fault detection without the requirement of known bounds on fault value or its derivative. It also enables rapid and fixed-time fault-tolerant control, offering predefined performance via defined funnel bounds on tracking error. Lyapunov stability analysis demonstrates the ultimate boundedness of estimation errors with the proposed observer and the fixed-time stability of the control system. 

A fault diagnosis and fault tolerant control scheme is proposed in \cite{kang2020sensor} for a leader-follower multiple manipulator system with sensor faults under a network communication topology. A learning observer is designed to estimate sensor faults in follower manipulators, and a distributed multiple manipulator synchronization controller is designed for each follower manipulator using fault estimation and a radial basis function (RBF) neural network approximator. The stability of the error system is analyzed using the Lyapunov stability theorem. Observer-based fault-tolerant synchronized control of multi-agent systems with faults is studied in \cite{deng2017distributed,zhu2017cooperative}. 

In \cite{sun2020fault} a novel adaptive descriptor observer and controller for FE and FTC of discrete-time nonlinear stochastic systems with multiple time-varying state delays and intermittent sensor and actuator faults is developed. The adaptive descriptor observer estimates the error dynamics and performs n-step FE. The active observer-based controller stabilizes the closed-loop system. A series of delay-dependent sufficient conditions are derived using the delay-dependent Lyapunov function and linear matrix inequalities. While in \cite{sun2020fault2} a fuzzy adaptive descriptor observer and controller is designed for FE and FTC of a discrete time Takagi–Sugeno (T–S) fuzzy stochastic systems with multiple time-varying state delays, intermittent sensor and actuator faults, nonlinear dynamics, and exogenous disturbances.

A fault diagnosis strategy for industrial robotic manipulators based on inverse dynamics-based feedback linearization is presented in \cite{incremona2019fault}. The strategy uses vision-based logic to detect sensor faults and a set of second-order sliding mode unknown input observers to detect and isolate actuator faults. The sliding-mode approach provides good stability, robustness, and fault estimation accuracy. A vision sensor is used to distinguish between sensor and actuator faults and to design a fault-tolerant control strategy in the case of sensor faults. The strategy has been verified and validated through simulation and experiments on an industrial robotic manipulator.

An integrated FD and FTC scheme for robotic manipulators with sensor faults is proposed in \cite{hong2019integrated}. The adaptive observer-based fault diagnosis scheme rapidly and accurately estimates constant and time-varying faults, and uses linear matrix inequality (LMI) method solves the design parameters. The fault-tolerant controller combines proportional-derivative and sliding mode control to track the desired trajectory using the back-stepping method.

Recent advances in sensor fault diagnosis in manipulators have enabled the development of more reliable and robust control systems for a variety of applications, such as manufacturing, robotic surgery, and space exploration. A robust fault detection and isolation for sensors used in robot manipulators is provided in \cite{paviglianiti2010robust}. A new hybrid robust FTC scheme is proposed in \cite{hagh2017hybrid}, switching between a robust H$\infty$ controller and a non-singular terminal sliding mode controller upon actuator fault detection. The system is first linearized using feedback linearization. A fuzzy-based switching system is used to switch controllers, and an adaptive joint unscented Kalman filter is used for fault detection and diagnosis. The proposed method simultaneously estimates system states and parameters. A Robust FTC for n-DOF serial hydraulic manipulators under matched and mismatched uncertainties and sensor faults is proposed in \cite{truong2021robust}. The system is modeled and a robust FE algorithm based on extended state observer is proposed which achieves force and position tracking stability.

\subsection{\textbf{Advancements in Sensor Technology}}

The utilization of sensor technology in recent times has been essential in augmenting fault tolerance and fault detection capabilities across diverse sectors. The integration of advanced sensors provides more accurate and real-time data, enabling better monitoring and detection of anomalies in systems \cite{shin2022micro}. A comprehensive review of the advancement of sensor technologies is presented in \cite{kalsoom2020}. These developments have significantly enhanced the ability to detect and diagnose faults in robotic manipulators \cite{milecki2023}. Here are several ways in which advancements in sensor technology contribute to improved fault tolerance and fault diagnosis:

\begin{itemize}
    \item Precision and Resolution: Modern sensors offer higher precision and resolution, allowing for finer measurements. This level of accuracy is critical for detecting subtle deviations in the robot's behavior, which might be indicative of early-stage faults \cite{nair2021emerging}.
    \item Multi-Sensor Fusion: The integration of multiple sensors, known as multi-sensor fusion, has become more prevalent. Combining data from different sensors, such as encoders, force/torque sensors, and vision systems, provides a more comprehensive view of the robot's state and its interactions with the environment \cite{alatise2020review}.
    \item Sensor Redundancy: Redundancy in sensor systems ensures that even if one sensor fails, others can continue to provide data for fault detection and control adaptation. This redundancy is particularly important for critical applications where safety and reliability are paramount \cite{zhang2020review}.
    \item Smart and Adaptive Sensors: Smart sensors equipped with built-in processing capabilities and adaptive functionalities can analyze data locally. This reduces the need for continuous data transmission to a central processing unit, minimizing communication delays and improving the efficiency of fault diagnosis \cite{ha2020machine}.
   \item Integration with Internet of Things (IoT) Platforms: Sensors are integral to IoT platforms, creating a network of interconnected devices. This interconnectedness allows for centralized monitoring, data analysis, and decision-making. IoT-based fault diagnosis systems leverage sensor data to provide a holistic view of the entire system \cite{shaikh2022, romeo2020}.
   
\end{itemize}

\subsection{\textbf{Integration of Sensor Data into Fault-Tolerant Control}}
The data collected by sensors serves as the foundation for FD and FTC strategies. The data set may consist of data collected from several sensors integrated into the system as well as feedback data obtained from various components of the system. These data set needs to be processed and analysed for decision-making.

A comprehensive review on FDD strategies within a unified data-processing framework, offering different data processing techniques on FDD in complex systems is discussed \cite{dai2013model}. A detailed review on Deep Transfer Learning Models for Industrial Fault Diagnosis Using Vibration and Acoustic Sensors Data is discussed in \cite{bhuiyan2023}. A review exploring Big Data Management Tools, Sensing and Computing Technologies, using sensor data, Visual Perception and Environment Mapping Algorithms in the Internet of Robotic Things is discussed in \cite{remote2022}.

Relevant review \cite{1s,2s,3s} utilizing sensor data for fault detection and diagnosis, suggest some key steps from data collection to fault diagnosis and reconfiguration. The involved key steps are discussed below.
\begin{itemize}
    \item Data Collection: Sensors continuously gather data regarding the manipulator's position, orientation, joint velocities, joint torques, and any external forces or torques acting on the system. The data is transmitted to the control system in real time.
    \item Data Analysis: The control system processes the sensor data to assess the system's state. This includes checking for inconsistencies or deviations that may indicate a fault. Techniques like signal processing, data fusion, and filtering are employed to enhance the quality of the data and extract relevant information.
    \item Fault Detection: Fault detection algorithms analyze the processed sensor data to identify anomalies. These algorithms may use predefined thresholds or machine learning techniques to detect deviations from the expected behavior. For instance, a sudden change in joint torque values could be indicative of a mechanical fault.
    \item Diagnosis: Once a fault is detected, the control system's diagnostic module determines the nature and location of the fault. This step is crucial in determining the appropriate response for fault tolerance.
    \item Control Adaptation: With knowledge of the fault, the control system adjusts its strategies to compensate for the fault's effects. Depending on the fault's severity, these adaptations may range from simple adjustments to more complex reconfiguration of control laws or trajectories.
    \item Recovery and Graceful Degradation: In cases where full recovery is not possible, the control system ensures graceful degradation, allowing the robotic manipulator to continue its operation at a reduced level of performance or functionality. This prevents a complete system shutdown and minimizes disruptions.
\end{itemize}

Overall, sensor-based FTC leverages advancements in sensor technology to detect and respond to faults in real time. By providing precise, real-time information about the robot's state, sensors enable control systems to enhance fault tolerance and system reliability, making them a critical component of modern robotic manipulators.

\section{Recent Advances in AI and Machine Learning}

In recent years, the fields of AI and ML have made significant contributions to the advancement of FTC, FD and FDI for complex systems and processes such as manufacturing, chemical processing, and robotics. These technologies have empowered robotic systems with the ability to detect and respond to faults in a more intelligent and adaptive manner.

In this section, we will explore the applications of AI and ML in FTC, focusing on the use of ML, in particular deep learning, for fault detection and diagnosis, the role of reinforcement learning (RL) in fault recovery strategies, and the integration of human-in-the-loop control systems with AI.

AI and ML technologies have transformed the landscape of FTC by providing sophisticated tools for fault detection, diagnosis, and recovery. The key applications of AI and ML in FTC include:

\begin{itemize}
    \item Anomaly Detection: AI algorithms, particularly those based on DL, have been employed for anomaly detection. By training models on normal robotic behavior, any deviation from the learned patterns can be flagged as an anomaly. This allows for the early detection of faults and malfunctions.
    \item Predictive Maintenance: ML models can predict when specific components of the robotic system are likely to fail based on historical data and environmental factors. This enables proactive maintenance to prevent faults and reduce downtime.
    \item Adaptive Control: AI and ML can adapt control strategies in real time based on the observed behavior of the robotic system. This adaptability ensures that the system continues to operate optimally in the presence of faults or disturbances.
    \item Fault Recovery: ML models, especially RL, enable the development of intelligent fault recovery strategies. By learning optimal control policies, robotic manipulators can autonomously recover from faults or deviations from desired behavior.
\end{itemize}

\subsection{Artificial Intelligence for Fault-Tolerant Control}

The field of robotics and AI, has the capacity to enhance and augment human capabilities, leading to increased productivity. Furthermore, these technologies are progressing beyond basic thinking and are now striving to emulate human-like cognitive capacities \cite{AI1}. AI approaches are extensively advanced technologies that find utility in the domains of fault-tolerant control and fault diagnostics for robotic systems. Fuzzy logic controllers (FLC), Bayesian reasoning techniques, and artificial neural network (ANN) models are among the AI technologies employed in fault-tolerant control and problem diagnostics \cite{AI2}.

As early as 1987, the integration of AI in FTC emerged when Selkäinaho and Halme proposed a real-time expert system for fault identification and localization in \cite{AI3}. In case of sensor failure, the system promptly substituted the measurement with a predictive model output signal, demonstrating success in a pilot test addressing physical flaws through AI. The authors in \cite{AI4} developed an AI model using adaptive fuzzy type-2 backstepping control (AFT2BC) method for robotic manipulators. AFT2BC successfully controlled a PUMA560 robotic arm in a MATLAB simulation, even when the robot's kinematic configuration was changed to simulate axis failure or load changes. The proposed control algorithm did not require prior knowledge of the robot's dynamic model, allowing it to operate effectively under both model uncertainty and external disturbances.

The work in \cite{AI5} proposed a FTC scheme using non-singular terminal synergetic control and interval type-2 fuzzy satin bowerbird optimization (IT2FSBO). Additionally, an adaptive augmented extended Kalman filter (A-AEKF) was presented in \cite{AI6} to detect, identify, and isolate actuator faults with noise resilience.

In \cite{AI7}, authors introduced a neuro-fuzzy robot fault-detection algorithm enabling control with an SRI \cite{AI8} controller despite sensor or axis actuator failures. The architecture utilized a multilayer perceptron trained with backpropagation and a FLC block for detection and FTC. The ANN input layer included 15 neurons, organized into 3 groups, assessing position, velocity, and acceleration for each axis. Output neurons generated positions and velocities for each axis, integrated with fuzzy logic block results. The algorithm successfully identified damaged axes, tested on a simulation of the ER5u robot in MATLAB. While in \cite{AI8}, the work proposed a FD system for a 5-DOF robotic arm using a neuro-fuzzy approach. The system comprises a MLP for FD and a fuzzy logic rule base for fault type and location identification. Simulation results demonstrate the effectiveness of the NF-based FD system.

The authors in \cite{AI9} proposed a novel FD and FTC method for robotic manipulators, combining an SVM-based neural adaptive high-order variable structure observer (ANHWSO) with an AI modelled adaptive modern fuzzy backstepping variable structure controller (AMFBVSC). The method demonstrated improved fault identification performance compared to traditional approaches. The work in \cite{li2019fault} introduced a dual neural network approach for FD and FTC in robotic manipulators, enabling control even when up to three axes fail simultaneously. This was the first study to demonstrate FTC capability in case multiple axes failure.

\subsection{Deep Learning for Fault Detection and Diagnosis}

The remarkable performance of DL in diverse fields like biomedical research, image recognition, natural language processing, and voice recognition has captivated researchers worldwide in recent years \cite{chebel2017, chen2021dynamic, pecht2009, raouf2022sensor, khalid2021real}. Research in robotic fault detection has witnessed significant advancements through DL techniques \cite{qiao2016advancing, dash2023deep, pu2020exploiting, adam2023multiple}. DL algorithms boast immense potential, capable of automatically extracting complex and intricate features from raw data, eliminating the need for manual feature engineering \cite{alam2020survey, lundervold2019overview, rohan2020rotate}. This powerful ability to discover hierarchical features within vast and multi-dimensional industrial data sets makes DL a highly promising tool for predictive maintenance solutions \cite{li2019deep, wang2018deep, duan2018deep, lee2023prognostics}.

Traditional data-driven approaches heavily relied on manual feature engineering, requiring signal processing expertise and significant human intervention. These techniques struggled with large data volumes and lacked real-time capabilities. DL models revolutionize this process by automatically extracting and selecting relevant features directly from raw data, eliminating the need for time-consuming and error-prone manual feature engineering. This enables them to learn complex patterns and representations from the data directly, leading to improved fault detection \cite{hernavs2018deep, fernandes2022machine, abid2021review}. DL surpasses traditional methods due to its ability to handle intricate patterns and non-linear interactions in robotic systems, along with end-to-end learning, facilitating the development of enhanced FDD. DL techniques incorporate different DL architectures, including Restricted Boltzmann Machine (RBM) \cite{kumar2023prognostics}, autoencoders (AEs) \cite{fathi2021predictive}, convolutional neural networks (CNNs) \cite{pan2021sensor, pan2022deep}, and recurrent neural networks (RNNs) \cite{wang2009recurrent}. These frameworks have been proven highly effective for FDD in robotic manipulators.

RBM is a generative stochastic neural network framework. offer a powerful unsupervised learning approach for fault detection. There aee two generative deep neural network (DNN) models based on the RBM; namely, the deep belief network (DBN), and deep Boltzmann machines (DBMs) \cite{zhang2018overview}. RBMs excel at uncovering hidden patterns in data, making them ideal for anomaly detection in complex systems. Trained on normal behavior data, they identify deviations as potential anomalies. This allows RBMs to act as pre-processors, extracting key features from high-dimensional sensor data or generating compressed representations. These features can then be fed into downstream models for further analysis and decision-making \cite{schmah2008generative, nayak2021}. While not as widely used as other DL frameworks in FTC/FDD, RBMs' unique capabilities offer significant potential for enhancing fault detection accuracy and reliability \cite{erfani2016high, mansouri2021deep}.

AEs are powerful unsupervised learning tools that efficiently learn data representations. They consist of two parts: an encoder that compresses the input data and a decoder that reconstructs it \cite{goodfellow2016deep, chen2023context, vincent2010stacked}. Autoencoders excel at compressing data while simultaneously capturing crucial information, making them valuable for extracting meaningful features from sensor data in FTC/FDD applications \cite{park2019fault, jiang2017wind,qian2022review}. This feature extraction capability, achieved through training on large datasets, allows the encoder to learn representations that serve as valuable inputs for subsequent models like fault classifiers or prognostics \cite{yang2022autoencoder}. Additionally, AEs can reduce the dimensionality of high-dimensional sensor data, simplifying processing and storage while preserving crucial information. This compression into a lower-dimensional latent space facilitates further analysis and model development \cite{peyron2021latent,fan2017autoencoder}.

CNN are primarily utilized for image-based fault detection, where visual data, such as images captured by cameras, serves as the input \cite{browne2003convolutional, oh2019convolutional}. These networks possess the remarkable ability to analyze images of robotic manipulator components and identify visual anomalies, like cracks or deformations, potentially signifying an underlying fault \cite{li2020rolling}. 

RNNs possess the remarkable ability to learn and exploit temporal dependencies in time-series data generated by sensors, making them uniquely suited for predicting and diagnosing faults \cite{koylu2020rnn}. By analyzing the evolving patterns in sensor data, RNNs can effectively detect anomalies such as changes in joint angles or torques, which might indicate mechanical faults \cite{wu2010robust}. These DL models achieve unprecedented levels of accuracy and precision in fault detection, surpassing traditional methods. Through extensive training on large datasets, RNNs learn to recognize both common and rare fault patterns, significantly enhancing the robustness of FTC systems. A comprehensive review on RNN application for FTC/FD in mechanical system is discussed in \cite{zhu2022}.

FTC/FDD using DL has proved to be a fast-growing field, with a lot of studies being conducted for robotic manipulators. Lee et al. \cite{lee2023prognostics} proposed an adaptive fault detection approach for servo-motors, enabling robustness under varying operating conditions. Adam et al. \cite{raouf2023transfer} developed a multi-fault diagnosis approach utilizing a CNN algorithm, paving the way for the detection of multiple concurrent faults. Rauf et al. \cite{raouf2023transfer} adopted a transfer learning-based DL framework for efficient fault detection in industrial robots. Zhou et al. \cite{zhou2022harmonic} successfully applied a DL model for diagnosing faults in harmonic reducers, demonstrating its effectiveness in identifying potential issues. Yin et al. \cite{yin2023} designed a dual-driven transfer network specifically for fault diagnosis in industrial robots, highlighting its potential for this application. These advancements pave the way for more reliable and robust robotic systems through efficient and accurate fault detection capabilities.


\subsection{Reinforcement Learning for Fault Recovery Strategies}

Reinforcement learning (RL), a powerful class of ML algorithms, can significantly enhance the accuracy of control parameter estimations and enable the creation of adaptive controllers that learn and adapt online in real-time \cite{hafner2011reinforcement, bu2020fuzzy, bu2019adaptive}. RL, with its inherent intelligence and learning abilities, interacts with its environment to discover the best strategy for maximizing its value function. Popular learning algorithms include policy iteration, value iteration, integral RL, and Q-learning \cite{hua2018new, szepesvari2022}.

RL has emerged as a potent tool for developing intelligent fault recovery strategies in robotic manipulators. In RL, an agent learns to interact with its environment to maximize a reward signal. The agent takes actions, observes the resulting state changes, and learns which actions lead to the best outcomes over time. A RL-integrated active fault-tolerant control framework for robotic manipulators with joint actuator faults is proposed in \cite{yan2022active}. This framework utilizes a neural network for fault diagnosis and RL for fault-tolerant control. When an actuator fault is detected, the RL controller generates compensation torques to maintain system safety and control performance. This approach avoids relying on precise system models, making it suitable for broader applications. The framework was evaluated on a 7-DOF Panda manipulator in the MuJoCo simulator, demonstrating its effectiveness.

Zhu et.al in \cite{zhu2023new}, propose a novel RL approach for fault-tolerant control of flexible multijoint robots. The method utilizes a model-free adaptive algorithm with parameter estimation and neural networks to achieve accurate fault identification and compensation while simultaneously reducing computational complexity. This is achieved through a specially designed critic-actor mechanism with an event-triggered parameter selection strategy. zhang et. al. in \cite{zhang2023dynamic} presented a novel active FTC framework for robot manipulators using learning algorithms. This approach, unlike conventional FTC methods, leverages dynamic learning theory and radial basis function networks to achieve accurate identification and learning of both uncertainties and actuator faults.

The work in \cite{sacchi2023sliding} proposes a FD and FTC scheme for redundant robot arms. It combines deep RL (DRL) and sliding mode observers. The DRL detects and isolates sensor faults, while the observers identify and compensate for actuator faults. This approach improves the safety and reliability of robotic systems. The work is simulated for a 7-DOF Franka Emika Panda robot arm using PyBullet environment. Liu et. al. in  \cite{liu2016} developed an adaptive FTC scheme for MIMO nonlinear discrete-time systems utilizing RL. The proposed approach significantly reduces learning parameters in the RL-based FTC policy via a fast-updating law, enhancing efficiency and reducing computational burden. Its effectiveness is demonstrated on a two-link planar robot manipulator, showcasing the strength of this approach in achieving robust and efficient FTC.

Based on literature \cite{liu2023reinforcement, qian2022development, siraskar2023} and capabilities of RL \cite{elguea2023review, }, it can be said that in the context of FTC, RL can be applied as follows:

\begin{itemize}
    \item Learning Optimal Control Policies: An RL agent can learn optimal control policies that help the robotic manipulator recover from specific types of faults. By training in simulated environments, the agent can explore various strategies for fault recovery and adapt its behavior in real time \cite{moos2022robust}.
    \item Adaptive Control: RL can be used for adaptive control, where the agent continually adjusts control parameters to adapt to changing conditions and faults. This dynamic control strategy allows the robot to maintain desired performance even in the presence of uncertainties \cite{barzegar2022deep}.
    \item Incorporating Prior Knowledge: RL models can incorporate prior knowledge about the robotic system's dynamics and constraints, enhancing their efficiency in fault recovery. For example, an RL agent can consider the robot's kinematics and dynamics to optimize control inputs. One notable advantage of RL is its ability to handle complex, dynamic, and partially observable environments, making it well-suited for real-world applications where faults may arise in unpredictable ways. RL-based fault recovery strategies enable robotic manipulators to adapt to changing conditions and recover gracefully from faults, reducing the impact of failures on operations \cite{liu2021deep}.
\end{itemize}

\section{Digital Twin Technology in Fault-Tolerant Control}

The confluence of DTT and FTC has revolutionized the reliability and performance landscape within the smart manufacturing industry \cite{he2021digital}. With Industries 4.0 and 5.0 heavily reliant on robotic manipulators, this section embarks on a thorough examination of DTT's significance in the context of FTC, both generally and specifically as applied to robotic systems. We will delve into DTT's pivotal role in predictive maintenance, preempting failures and bolstering FTC strategies through real-time simulations \cite{frej2022intelligent}.

DTT creates a virtual twin of a physical system replicating its structure, behavior, and performance. Real-time data synchronizes the twins, enabling in-depth monitoring, predictive analysis, and virtual testing of control strategies \cite{juarez2021digital, batty2018digital
}. DTT bridges the physical-digital gap, providing a live reflection of the robot's state and performance. This real-time model empowers FTC by predicting faults, enabling proactive interventions, and offering a virtual sandbox for testing fault tolerance strategies. DTT's high-fidelity representation enhances system monitoring and performance analysis \cite{huang2021survey}.

The work in \cite{wu2022low} proposes a DT model to control a robotic arm's positioning error. The DT model, encompassing physical, data, model, functional, and application layers, utilizes a low-cost attitude sensor for real-time feedback. This allows the DT to detect and mitigate positioning errors by adjusting motor angles. While in \cite{wu2023digital} a DT-enabled system for 3D positioning and error compensation in robotic arms is designed. A virtual sensor, modeled geometrically, exchanges information with the physical sensor, enabling real-time pose comparison. Closed-loop alignment between physical data and virtual outputs dynamically adjusts arm joints to minimize positioning errors. This information mutuality drastically reduces angle calculation needs, leading to an impressive 81.23\% error reduction across diverse positions.

Predictive maintenance has emerged as a vital facet of fault tolerance, allowing proactive actions to prevent failures in robotic manipulators. Digital twins play a pivotal role in enabling predictive maintenance by continuously monitoring the state of the physical system and assessing its condition. A detailed review titled "Deep digital twins for detection, diagnostics and prognostics" is presented in \cite{booyse2020deep}. This study centers on developing a deep digital twin (DDT) using deep generative models, learning the healthy data distribution directly from operational data at an asset's inception. Unlike traditional methods, DDT doesn't depend on historical failure data for estimating asset health, enabling predictive fault maintenance.

Based on the literature \cite{liu2018role, aivaliotis2019use, you2022advances}, it can be said that the process of leveraging DT for predictive maintenance encompasses the following key elements:
\begin{itemize}
    \item Data Synchronization: Sensors placed on the physical robotic manipulator collect data regarding its state, including joint angles, velocities, temperatures, and other relevant parameters. This data is transmitted in real time to the digital twin.
    \item Analysis and Comparison: The digital twin processes and analyzes the received data to create a dynamic representation of the physical system. Deviations from the expected behavior are identified by comparing the real-time data with the model's predictions.
    \item Early Fault Detection: By recognizing early signs of anomalies or performance degradation, digital twins can flag potential issues that may lead to failures. For instance, if a joint's performance deteriorates beyond a certain threshold, the digital twin can signal that maintenance is required.
    \item Predictive Models: Machine learning algorithms can be integrated into the digital twin to create predictive models. These models use historical data to forecast when specific components of the robotic manipulator are likely to fail or require maintenance.
    \item Proactive Actions: Armed with predictive insights from the digital twin, maintenance actions can be initiated in a timely and precise manner. This preventive maintenance reduces the risk of unexpected failures and minimizes downtime.
\end{itemize}

Predictive maintenance through digital twins offers a data-driven approach to ensuring the reliability of robotic manipulators. It shifts the paradigm from reactive maintenance, which addresses failures after they occur, to a proactive strategy that identifies issues before they can disrupt operations. Xu et. al. in \cite{xu2019digital} proposed a two-phase digital-twin-assisted fault diagnosis method using deep transfer learning. A ML-based architecture for sensor FD and FDI using DT is discussed in \cite{darvishi2022machine}. 

DT technology proves highly transformative in FTC by providing the ability to establish a virtual, real-time simulation environment for robotic manipulators. This virtual space replicates the physical system, enabling risk-free testing, optimization, and the formulation of fault tolerance strategies. The process of using DT for real-time simulation involves several key components:
\begin{itemize}
    \item Dynamic Simulation: The DT uses real-time data from the physical robotic manipulator to update its dynamic simulation. This simulation models the manipulator's kinematics, dynamics, and behavior, offering a real-time representation of the system's operation \cite{ganguli2020digital}.
    \item Control Algorithm Testing: Control strategies and fault tolerance algorithms can be tested in the virtual environment using the digital twin. This testing allows for the evaluation of various control parameters and strategies to optimize fault tolerance \cite{he2019data}.
    \item Scenario Testing: DT facilitate the creation of diverse scenarios that mimic real-world conditions and potential fault scenarios. These scenarios can be used to assess how the robotic manipulator responds to various challenges and disturbances \cite{becue2020new}.
    \item Training and Optimization: ML techniques, including reinforcement learning, can be employed to optimize control strategies and fault tolerance responses. The DT provides a safe space for training and testing these algorithms \cite{xia2021intelligent}.
    \item Real-Time Response Analysis: During real-time simulation, the DT continuously analyzes the robotic manipulator's response to various inputs and disturbances. This analysis aids in refining control strategies and enhancing fault tolerance \cite{darvishi2021real}.
\end{itemize}

Digital twins provide a risk-free and efficient method for real-time simulation to refine fault tolerance strategies. This approach enables comprehensive testing and optimization in a controlled virtual environment, minimizing the need for costly physical experiments and reducing risks associated with real-world faults. They facilitate predictive maintenance, preventing failures proactively, and serve as a versatile platform for testing fault tolerance strategies. Integrating digital twins into robotic manipulator systems ensures enhanced fault tolerance, reduced downtime, and improved overall performance, establishing them as vital components in modern robotic manipulator control and maintenance.

\section{Integration of AI, ML, and Digital Twin }

The seamless integration of AI, ML, and DTT in FD/FTC for robotic manipulators represents a convergence that revolutionizes how these systems operate, anticipate, and adapt to faults. This section explores the synergies between these three technologies in FTC, presents case studies showcasing their successful integration, and discusses the benefits and challenges associated with this holistic approach.

\begin{figure*}[t] 
\centering
{\includegraphics[width= 17 cm]{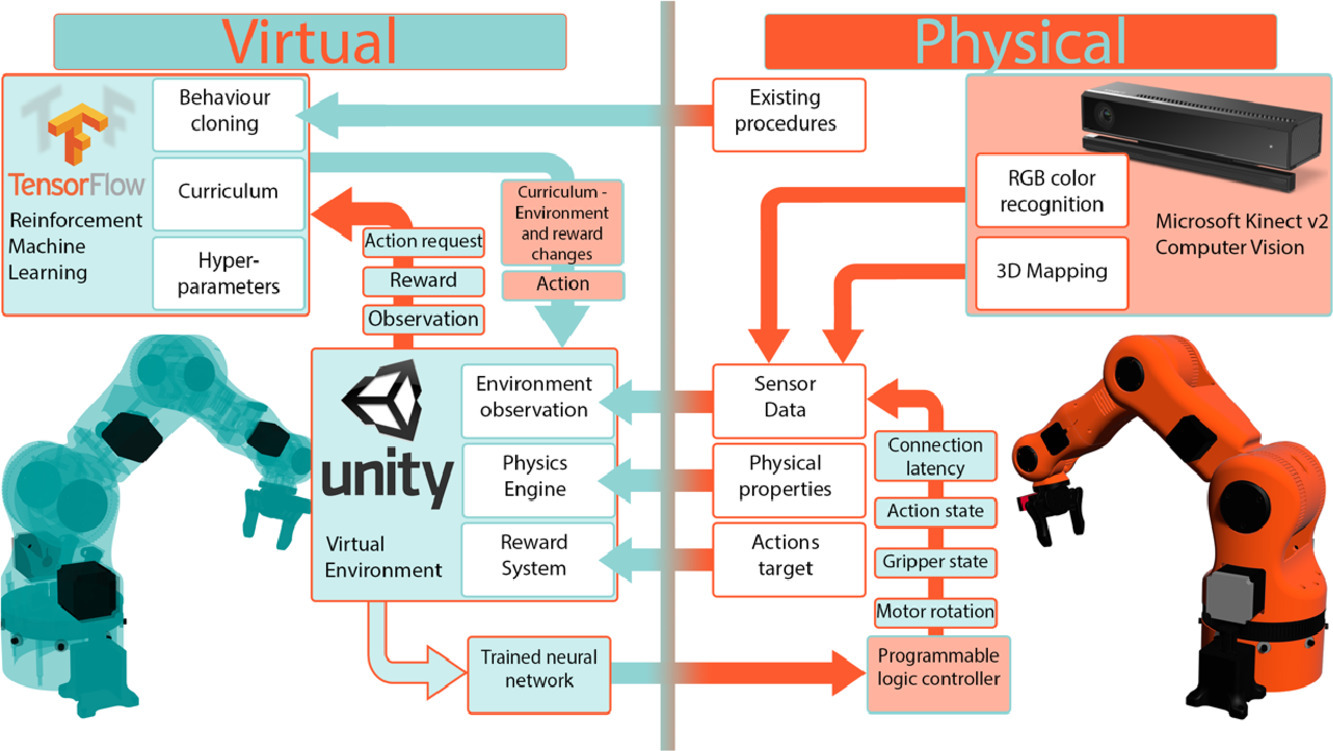}}
\caption{A robotic arm framework integrating RL and DT \cite{matulis2021robot}}
\label{RT_DT}  
\end{figure*}

\subsection{\textbf{Synergies Between AI, ML, and DTT in Fault-Tolerant Control}}

The integration of AI, ML, and DTT creates a powerful synergy that enhances the fault tolerance of robotic manipulators.
\begin{itemize}
    \item AI-Powered Digital Twins: AI-enhanced digital twins have the ability to adapt, learn, and predict. These digital replicas of the physical system can incorporate AI algorithms to monitor and analyze real-time data from the robot, identifying anomalies, predicting faults, and simulating fault scenarios \cite{2020digital, huang2021survey}. For example, a digital twin equipped with AI can analyze historical data and predict when a robot's joint might experience excessive wear, prompting preventive maintenance.
    \item Machine Learning for Anomaly Detection: ML algorithms, integrated into the digital twin, can learn patterns of normal system behavior. When deviations from these patterns occur, the digital twin can flag anomalies. These anomalies might signal incipient faults, prompting early intervention \cite{ademujimi2022digital}.
    \item Real-Time Control Optimization: AI and ML can optimize control strategies in real time based on data from the digital twin and physical robot \cite{rathore2021role}. This adaptability enables the robot to continue performing optimally in the presence of faults, such as motor failures or environmental disturbances.
    \item Fault Recovery Strategies: RL algorithms can be employed within the digital twin to train and optimize fault recovery strategies \cite{kim2024digital}. This learning process is carried out in a simulated environment, where the digital twin emulates the real robot's behavior. As a result, the robot can autonomously recover from various faults more effectively.
\end{itemize}

\subsection{\textbf{Benefits and Challenges Associated with This Integrated Approach}}
While the integration of AI, ML, and DTT in FTC offers numerous benefits, it also presents challenges:

\textbf{Benefits}:
\begin{itemize}
    \item Enhanced Fault Tolerance: Integration enables better fault detection, prediction, and recovery. This results in higher system reliability and uptime.
    \item Proactive Maintenance: Predictive maintenance through the analysis of historical data and real-time monitoring reduces the risk of unexpected failures, leading to cost savings and operational efficiency.
    \item Optimized Control Strategies: Real-time optimization of control strategies, driven by AI and ML, ensures optimal performance, even in the presence of faults or disturbances.
    \item Autonomous Recovery: RL-powered fault recovery allows robotic manipulators to autonomously respond to faults, minimizing the need for human intervention.
    \item Simulation for Testing: The digital twin provides a safe environment for testing fault tolerance strategies and fine-tuning control parameters without risking physical damage to the robot.

\end{itemize}

\textbf{Challenges}:
\begin{itemize}
    \item Complex Integration: Integrating AI, ML, and DTT can be complex and requires expertise in multiple domains.
    \item Data Handling: Managing vast amounts of data generated by sensors and digital twins can be challenging. Efficient data storage, processing, and analysis are necessary.
    \item Training and Calibration: Training AI and ML models, as well as calibrating digital twins, requires time and effort. The accuracy of these models depends on the quality of data and the realism of the digital twin.
    \item Cost: Implementing and maintaining integrated systems can be expensive. Organizations must weigh the costs against the benefits of improved fault tolerance and performance.
    \item Cybersecurity: Protecting the digital twin and the associated AI and ML systems from cyber threats is crucial. The integration of these technologies can introduce vulnerabilities that need to be addressed.
\end{itemize}


\section{ROBOTIC MANIPULATOR APPLICATION - USE CASES}

In this section, we will delve into detailed use case application that will exemplify real-world applications of FTC for robotic manipulators. These case studies highlight the pivotal role of AI, ML, and DT technologies in enhancing FDD/FTC.


\begin{enumerate}
    \item{\textbf{Automotive Manufacturing:}}
In an automotive manufacturing facility, robotic manipulators are responsible for tasks like welding and painting. A digital twin, equipped with AI and ML algorithms, continuously monitors the performance of these robots. AI is utilized for predictive maintenance, analyzing historical data and real-time sensor information to predict when components are likely to fail. ML, integrated with the DT, enables anomaly detection by learning normal robot behavior.

By leveraging AI, ML, and DT technologies, the automotive manufacturer achieved a significant reduction in unplanned downtime. Predictive maintenance allowed them to replace components before they failed, minimizing production interruptions. Anomaly detection led to early identification of issues, preventing defects in the final product. The integrated approach resulted in cost savings and improved product quality.

Implementing predictive maintenance through digital twins requires a robust data infrastructure to collect and analyze vast amounts of sensor data. Additionally, the models need to be continuously updated to adapt to changing conditions and data patterns.


    {\item{\textbf{Medical Robotics:}}}
In the field of medical robotics, DT play a vital role in surgical procedures. A DT combined with AI and ML, constantly monitors the robot's performance during surgery. AI algorithms optimize the robotic system's movements in real time to enhance precision and safety. The digital twin simulates various fault scenarios to allow the robot to practice fault recovery strategies.

The integration of AI, ML, and DT in medical robotics significantly improved the precision and safety of surgical procedures. The real-time optimization of robot movements reduced the risk of errors. The simulation of fault scenarios enabled the robot to respond effectively to unexpected events during surgery, ensuring patient safety.

The implementation of AI and ML in real-time surgical robotics requires rigorous testing and validation to ensure patient safety. It is essential to simulate a wide range of fault scenarios to prepare the robotic system for unexpected events during surgery.


    {\item{\textbf{Space Exploration:}}}
In space exploration missions, robotic manipulators are exposed to extreme environmental conditions. Digital twin technology, enhanced with AI and ML, continuously monitors the robot's performance. AI is used to predict potential faults based on environmental data, while ML models are trained to adapt the robot's behavior to changing conditions through RL. The DT simulates fault scenarios, enabling the robot to practice fault recovery strategies.

The integration of AI, ML, and DT technologies in space robotics significantly increased the reliability and adaptability of robotic systems. Predictive maintenance allowed for early intervention to prevent failures in the harsh space environment. The autonomous fault recovery strategies, learned through RL, ensured the robot could continue its mission even in the presence of unforeseen challenges.

Space exploration requires robust FTC systems, and the integration of AI, ML, and DT technologies must be rigorously tested to handle the unique conditions of space missions. Regular updates to the models and simulations are essential to adapt to the dynamic space environment.


\item{\textbf{Industrial Logistics:}}
In industrial logistics, robotic manipulators are used for tasks like sorting and packaging. A digital twin, coupled with AI and ML, continuously monitors the performance of these robots. AI-driven predictive maintenance ensures that components are replaced before they fail. ML algorithms integrated with the DT are responsible for detecting anomalies in robot behavior.

The implementation of AI, ML, and DT in industrial logistics led to a significant reduction in operational disruptions. Predictive maintenance prevented unexpected failures, reducing downtime in critical sorting and packaging processes. Anomaly detection ensured that defective products were identified early, preserving product quality and customer satisfaction.

Real-time data analysis and proactive maintenance through DT require robust infrastructure and ongoing model calibration. The integration of AI and ML must consider the unique requirements and operational conditions of industrial logistics.


\item{\textbf{Agriculture Automation:}}
In agriculture automation, robotic manipulators are employed for tasks like planting and harvesting. A DT, enhanced with AI and ML, continuously assesses the robot's performance. AI-driven predictive maintenance analyzes historical data and real-time sensor information to predict when components may fail. ML algorithms integrated with the digital twin enable anomaly detection by learning normal robot behavior.

The integration of AI, ML, and DT in agriculture automation resulted in increased efficiency and reduced maintenance costs. Predictive maintenance allowed for timely component replacements, minimizing disruptions during planting and harvesting seasons. Anomaly detection identified issues early, preventing crop damage and improving yield.

Implementing predictive maintenance in agriculture automation requires consideration of environmental factors and seasonal variations. Models must be adaptable to changing conditions and calibrated accordingly.

    \end{enumerate}

These use cases illustrate the significant impact of integrating AI, ML, and DT technologies in FTC for robotic manipulators across diverse industries. The outcomes demonstrate improved fault tolerance, reduced downtime, increased operational efficiency, and enhanced product quality. The lessons learned emphasize the importance of robust data infrastructure, rigorous testing, and adaptability to dynamic conditions when implementing these technologies in FTC systems.

\section{Challenges and Future Directions}
As the field of FTC for robotic manipulators continues to evolve, it faces a set of challenges and opportunities that shape its trajectory. In this section, we will examine the current challenges, discuss potential solutions and strategies, and outline emerging trends and future directions.

\subsection{Current Challenges in Implementing FTC for Robotic Manipulators}

\begin{enumerate}
    \item Complexity of Robotic Systems: Modern robotic manipulators are increasingly complex, with numerous components and sensors. Managing and monitoring these systems in real time poses significant challenges, as it demands robust fault detection, diagnosis, and recovery mechanisms.
    \item Data Management: The vast amount of data generated by sensors and digital twins can overwhelm traditional data storage and processing capabilities. Efficient data management and analysis are essential for effective FTC implementation.
    \item Modeling Accuracy: Developing accurate models for digital twins is a critical aspect of FTC. Modeling errors can lead to false alarms or undetected faults. Achieving high-fidelity models that represent the dynamics and behavior of robotic manipulators remains a challenge.
    \item Integration of Multiple Technologies: Integrating AI, ML, DT technologies requires a multidisciplinary approach. Ensuring seamless communication and compatibility among these technologies can be complex and challenging.
    \item Cybersecurity: The integration of DT and AI introduces cybersecurity concerns. Protecting the digital twin and associated AI systems from cyber threats is crucial to prevent unauthorized access and tampering with control algorithms.

\end{enumerate}

\subsection{Potential Solutions and Strategies to Address Challenges}

\begin{enumerate}
    \item Advanced Data Analytics: Implementing advanced data analytics, including big data and ML techniques, can assist in handling the vast amount of data generated by sensors. This can improve fault detection, diagnosis, and predictive maintenance.
    \item Improved Sensor Technology: The development of more advanced and reliable sensor technology can enhance the quality and quantity of data available for FTC. High-precision sensors with redundancy options can contribute to robust fault detection and diagnosis.
    \item Modeling Advancements: Continued research into modeling techniques can lead to more accurate digital twins. Utilizing physics-based models in conjunction with data-driven approaches can improve the prediction and diagnosis of faults.
    \item Standardization: Developing industry standards for the integration of AI, ML, and DT technologies in FTC can ensure compatibility and interoperability between different systems and components.
    \item Education and Training: Providing training and education to engineers and practitioners in the field of FTC is essential. Building expertise in AI, ML, and DT technologies will facilitate the implementation of these tools in practice.
\end{enumerate}

\subsection{Emerging Trends and Future Directions}

\begin{enumerate}
    \item Explainable AI (XAI): As AI plays a more prominent role in FTC, the need for XAI becomes critical. Developing AI systems that can explain their reasoning and decision-making processes will enhance transparency and trust in fault tolerance mechanisms.
    \item Edge Computing: Implementing edge computing in robotic manipulators can reduce latency and enhance real-time decision-making. AI and ML models can be deployed on the robot itself, allowing for faster fault detection and control adaptation.
    \item Quantum Computing: Quantum computing has the potential to revolutionize fault tolerance by rapidly solving complex optimization and decision-making problems. As quantum computing technology matures, it may find applications in FTC for robotic manipulators.
    \item Distributed and Multi-Agent Systems: The use of distributed and multi-agent systems for robotic manipulators can improve fault tolerance. Multiple robots can collaborate and adapt to faults in a coordinated manner, ensuring uninterrupted operation.
    \item Human-Machine Collaboration: Integrating human expertise with AI systems is an emerging trend. Human operators can work in collaboration with AI-powered control systems, providing insights and decision-making in complex fault scenarios.
    \item Adaptive Learning and Reinforcement Learning: Advancements in adaptive learning techniques and reinforcement learning algorithms can lead to more efficient and adaptive fault recovery strategies. These technologies will be critical for achieving autonomy in fault-tolerant robotic systems.
    \item Cross-Domain Transferability: Transferring knowledge and strategies from one domain to another is an emerging trend. Lessons learned from fault tolerance in one industry, such as manufacturing, can be applied to another, like healthcare robotics, with appropriate adaptations.
    \item Sustainability and Energy Efficiency: Future directions in FTC will also consider sustainability and energy efficiency. Optimizing control strategies to minimize energy consumption while maintaining fault tolerance is of growing importance.
\end{enumerate}

In summary, the challenges, and opportunities in the field of FTC for robotic manipulators are closely intertwined. Addressing the current challenges requires a combination of advanced technologies, interdisciplinary collaboration, and a focus on education and training. The emerging trends and future directions highlight the continued evolution of fault-tolerance control, with a growing emphasis on transparency, autonomy, and adaptability in robotic systems. As the field progresses, it will play an increasingly integral role in ensuring the reliability, safety, and performance of robotic manipulators across various industries.

\section{Conclusion}

This article has provided a comprehensive exploration of FTC for robotic manipulators, shedding light on the remarkable advancements driven by AI, ML and DT technologies. These technologies have ushered in a new era of fault tolerance, offering enhanced reliability, safety, and adaptability to robotic systems across various industries. Let's summarize the key takeaways from this discussion.

First and foremost, FTC plays a pivotal role in ensuring the uninterrupted operation of robotic manipulators. The ability to detect, diagnose, and recover from faults is essential in industries where precision, efficiency, and safety are paramount. Whether in automotive manufacturing, healthcare robotics, space exploration, industrial logistics, or agriculture automation, FTC's impact is profound.

The integration of AI, ML, and DT technologies is a game-changer in the field of FTC. AI-powered DT, combined withML, provide real-time monitoring, predictive maintenance, and anomaly detection capabilities. These technologies enable robotic manipulators to adapt to changing conditions, recover from faults autonomously, and reduce unplanned downtime. The result is not only cost savings but also improved product quality and operational efficiency.

The use cases presented in this article demonstrate the real-world applications of these integrated technologies. From automotive manufacturing to space exploration, the outcomes are clear: enhanced fault tolerance, reduced downtime, and improved safety. These examples also underscore the importance of proactive maintenance, precise fault detection, and autonomous recovery strategies.

Looking ahead, the field of FTC is poised for continued growth and significance. Emerging trends such as Explainable AI (XAI), edge computing, quantum computing, and human-machine collaboration promise to further enhance fault tolerance in robotic systems. Cross-domain transferability of knowledge and strategies will allow industries to benefit from one another's experiences.

Furthermore, the emphasis on sustainability and energy efficiency aligns with the broader goals of responsible technological development. Optimizing control strategies to minimize energy consumption while maintaining fault tolerance is crucial, especially in an era of increasing environmental consciousness.

In Conclusion, the synergy between AI, ML, and DT technologies is a revolutionary transformation in FDD/FTC for robotic manipulators. The continued integration of these technologies, coupled with ongoing research and innovation, will shape the future of robotics, ensuring that these systems not only perform with exceptional precision but also do so with unparalleled reliability, adaptability, and resilience in the face of unexpected challenges. As industries continue to rely on robotic manipulators for various applications, the growth and significance of FDD/FTC will remain at the forefront of technological progress.

\appendices

\ifCLASSOPTIONcaptionsoff
  \newpage
\fi

\begin{IEEEbiography}
[{\includegraphics[width=1in,height=1.25in,clip,keepaspectratio]{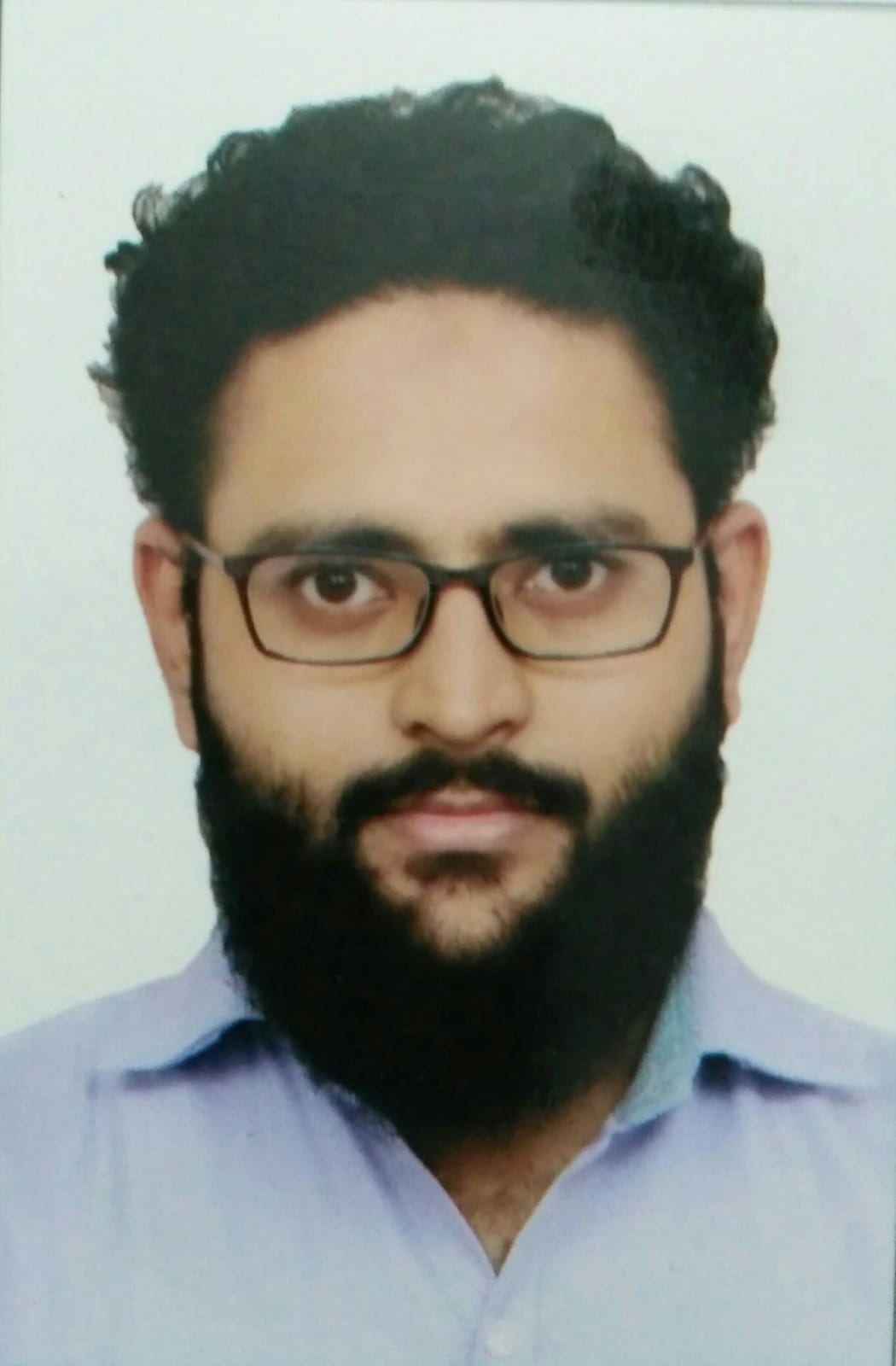}}]{Md Muzakkir Quamar,}
\
is currently pursuing his PhD in controls and Instrumentation Engineering Department at KFUPM. He obtained his MS degree in  controls and Instrumentation Engineering from KFUPM in 2022. He is the recipient of full-time scholarship for MS and PhD at KFUPM. His research Interest include nonlinear system, robotics, multi-agent system, optimal and robust control.
\end{IEEEbiography}

\begin{IEEEbiography}
[{\includegraphics[width=1in,height=1.25in,clip,keepaspectratio]{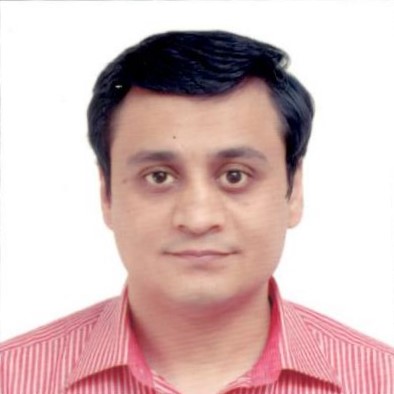}}]{Ali Nasir,}
\
is currently working as an assistant professor in the Control and Instrumentation Engineering Department at KFUPM. He is an affiliate of the Interdisciplinary Research Center for Intelligent Manufacturing and Robotics at KFUPM. He is also a guest affiliate of the Interdisciplinary Research Center for Aviation and Space Exploration. He obtained his PhD degree in Aerospace Engineering along with M.Sc. degrees in Electrical Engineering and Aerospace Engineering from the University of Michigan, Ann Arbor, USA. He received his B.Sc. in Electrical Engineering from University of Engineering and Technology, Taxila. He is the recipient of Fulbright scholarship for an MS leading to a PhD. He is currently working on fault-tolerant control and decision-making for industrial robots. His research interests also include approximate dynamic programming, nonlinear control, and estimation.
\end{IEEEbiography}

\nocite{*} 
\bibliographystyle{IEEEtran}
\bibliography{IEEEabrv,Main}

\end{document}